\documentclass[journal]{IEEEtran}

\usepackage{float}
\usepackage{ifpdf}
\usepackage{cite}
\usepackage[pdftex]{graphicx}
\usepackage{amsmath,amssymb,amsfonts}
\usepackage{algorithmic}
\usepackage{array}
\usepackage{fixltx2e}
\usepackage{url}
\usepackage{stfloats}
\usepackage{multirow}
\usepackage{multicol}
\usepackage{textcomp}
\usepackage{colortbl}
\usepackage{booktabs}
\usepackage{cite}
\usepackage[table]{xcolor}
\usepackage{amsmath}
\usepackage{bm}

\usepackage{makecell}
\usepackage{longtable}
\usepackage{graphicx}

\usepackage{subfigure}
\usepackage{subfloat}
\usepackage{cleveref}

\ifCLASSINFOpdf
\else
\fi

\begin{document}

\title{CodeEnhance: A Codebook-Driven Approach for Low-Light Image Enhancement}

\author{Xu Wu, XianXu Hou, Zhihui Lai$^{*}$, Jie Zhou, Ya-nan Zhang, Witold Pedrycz, Linlin Shen
\thanks{This work was accepted by Engineering Applications of Artificial Intelligence on May 7, 2025. Corresponding author: Zhihui Lai}

\thanks{X. Wu, Z. Lai, Ya-nan Zhang and L. Shen are with the Computer Vision Institute, College of Computer Science and Software Engineering, Shenzhen University, Shenzhen 518060, China, Shenzhen Institute of Artificial Intelligence and Robotics for Society, Shenzhen 518060, China, and Guangdong Key Laboratory of Intelligent Information Processing, Shenzhen University, Shenzhen 518060, China(e-mail: csxunwu@gmail.com, lai\_zhi\_hui@163.com; zyn962464@gmail.com; llshen@szu.edu.cn).}
\thanks{X. Hou is School of AI and Advanced Computing, Xi’an Jiaotong-Liverpool University, China (hxianxu@gmail.com).}
\thanks{J. Zhou is National Engineering Laboratory for Big Data System Computing Technology, Shenzhen University, and SZU Branch, Shenzhen Institute of Artificial Intelligence and Robotics for Society, Shenzhen, Guangdong 518060, China (e-mail: jie\_jpu@163.com).}
\thanks{W. Pedrycz is the Department of Electrical \& Computer Engineering, University of Alberta, University of Alberta, Canada (wpedrycz@ualberta.ca).}
}

\markboth{Journal of \LaTeX\ Class Files,~Vol.~14, No.~8, August~2015}%
{Shell \MakeLowercase{\textit{et al.}}: Bare Demo of IEEEtran.cls for IEEE Journals}

\maketitle

\begin{abstract}
Low-light image enhancement (LLIE) aims to improve low-illumination images. However, existing methods face two challenges: (1) uncertainty in restoration from diverse brightness degradations; (2) loss of texture and color information caused by noise suppression and light enhancement. In this paper, we propose a novel enhancement approach, CodeEnhance, by leveraging {discrete codebook priors} and image refinement to address these challenges. In particular, we reframe LLIE as learning an \textbf{image-to-code} mapping from low-light images to discrete codebook, which has been learned from high-quality images. To enhance this process, a Semantic Embedding Module (SEM) is introduced to integrate semantic information with low-level features, and a Codebook Shift (CS) mechanism, designed to adapt the pre-learned codebook to better suit the distinct characteristics of our low-light dataset. Additionally, we present an Interactive Feature Transformation (IFT) module to refine texture and color information during image reconstruction, allowing for interactive enhancement based on user preferences. Extensive experiments on both real-world and synthetic benchmarks demonstrate that the incorporation of prior knowledge and controllable information transfer significantly enhances LLIE performance in terms of quality and fidelity. The proposed CodeEnhance exhibits superior robustness to various degradations, including uneven illumination, noise, and color distortion. {Our project page is https://github.com/csxuwu/CodeEnhance.}
\end{abstract}


\IEEEpeerreviewmaketitle

\section{Introduction}
Suffering from low illumination intensity, diverse light sources, and color distortion issues, Low-Light (LL) images usually hinder visual perception and degrade the performance of downstream tasks \cite{ExDark}. To address these issues, the LLIE methods are proposed to obtain High-Quality (HQ) images from LL images \cite{LLNet}.

\begin{figure}[!t]
    \centering
    \includegraphics[width=0.48 \textwidth]{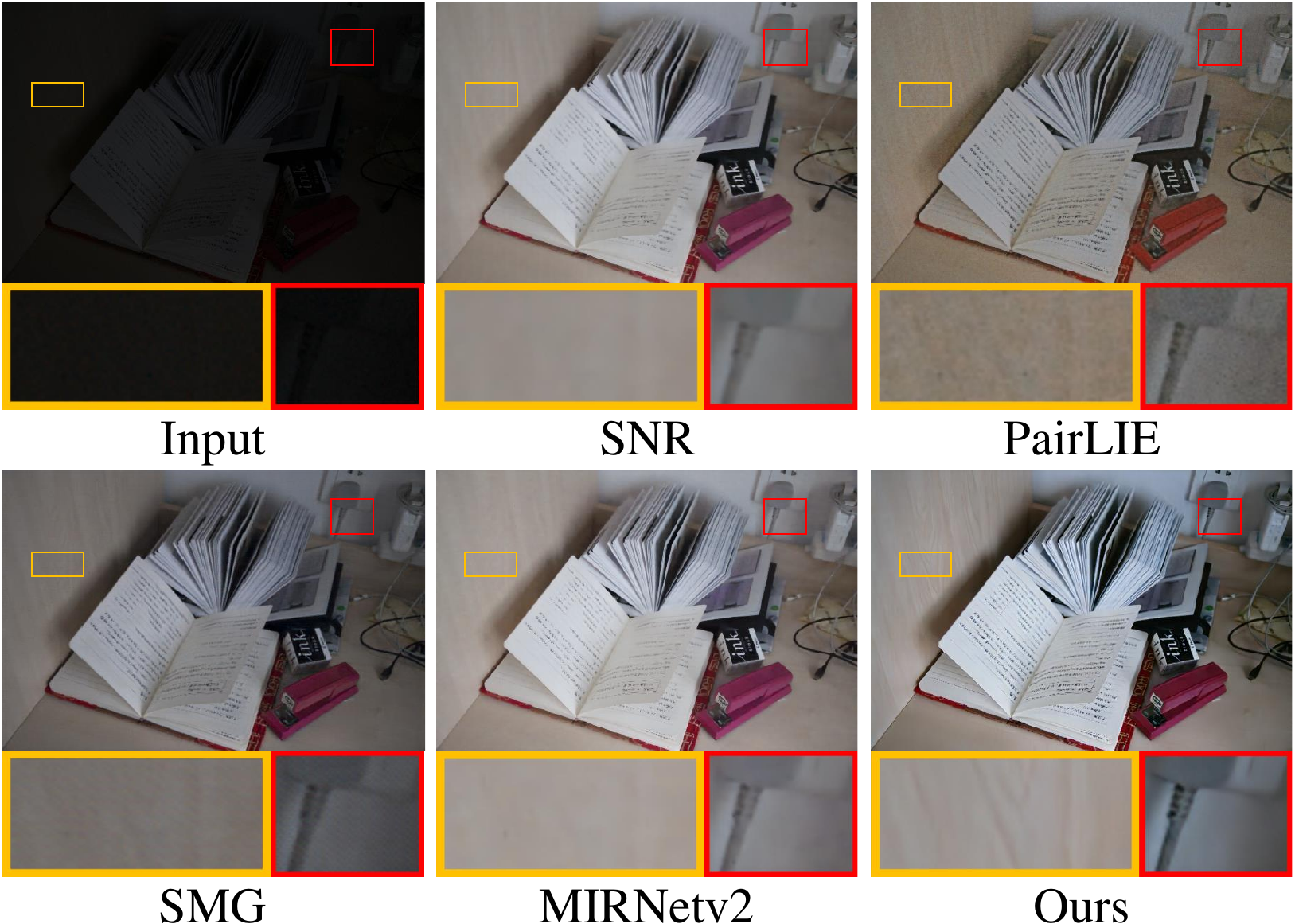}
    \caption{Visual comparison on the LOL \cite{RetinexNet} dataset. It can be found that the proposed CodeEnhance outperforms other methods (SNR \cite{SNR}, SMG \cite{SMG}, and MIRNetv2 \cite{MIRNetv2}) in terms of light enhancement, texture maintenance, noise suppression, and color restoration.} 
    \vspace{-4mm}
    \label{fig:example}
\end{figure}

Classic LLIE methods based on histogram equalization \cite{AHE} and retinex theories \cite{SSR}\cite{Land1971LightnessAR}, are effective in enhancing image brightness. {However, these methods neglect image content and color, resulting in a distortion of the output image \cite{MEF}\cite{AGLLNet}. }
Recently, deep learning-based methods have been leveraged due to their powerful expressiveness. The popular methods are mainly based on adversarial training \cite{EnlightGAN} or image-to-image \cite{TCI2}\cite{TCI5} framework, 
both of which aim to learn the mapping between LL and HQ images. Nevertheless, existing deep learning-based methods encounter limited generalization in real night scenes that contain complex light sources and intricate illumination. 
% Concretely, 
% (1) The restoration process is inherently uncertain due to the challenge of addressing diverse brightness degradations, which complicates accurate reconstruction.
% (2) Loss of texture and color often occurs due to the challenging balance between noise suppression and light enhancement in LLIE, as illustrated in Fig. \ref{fig:example}.
{Concretely, 
\textit{(1) Uncertainty in LLIE:}
Real-world nighttime scenes exhibit a broad spectrum of brightness degradations, making it inherently challenging to achieve accurate reconstruction. This diversity introduces uncertainty into the restoration process, as conventional methods often cannot capture the complexity of varying illumination conditions.
\textit{(2) Tradeoffs in LLIE:}
In low-light scenarios, removing noise often comes at the expense of texture and color information. As shown in Fig. \ref{fig:example}, current methods frequently over-smooth or distort the enhanced image, underscoring the challenge of balancing noise suppression with the need to preserve fine details and natural color.}

To overcome these challenges, we propose a novel approach named CodeEnhance by 
feature matching with quantized priors and image refinement.
Previous methods usually learn an LL-HQ image mapping, which typically involves a vast parameter space that {introduces uncertainty for the learning \cite{Codeformer}.} 
The key to reducing the uncertainty is to shrink the parameter space. Therefore, we reframe the LLIE task as an \textbf{image-to-code} paradigm by extracting image features and converting them to the codebook indices, followed by the pretrained decoder to generate the output. This approach efficiently reduces the parameter space, which alleviates uncertainties in the restoration process and improves robustness against various brightness degradations.
Besides, we use the shallow features to compensate for texture information in the reconstruction process. We further introduce HQ images to guide accurate color enhancement in the output, thereby enhancing visual perception.

Specifically, a Semantic Embedding Module (SEM) is developed to integrate semantic information and low-level features in the encoder of CodeEnhance. 
The SEM is crucial for bridging the semantic gap between the encoder output and the codebook, leading to effective feature matching. 
Moreover, to address distribution shifts across datasets, we introduce a Codebook Shift (CS) mechanism. 
This mechanism is designed to adapt the pre-learned codebook from one dataset to better suit the distinct characteristics of our LLIE dataset. It ensures distribution consistency and emphasizes relevant priors in feature matching.
To improve the restoration of texture and color, we design an Interactive Feature Transformation (IFT) module to fine-tune the texture, color, and brightness of the output image. 
The IFT consists of two main components: Texture Feature Transformation (TFT) module and Controllable Perceptual Transformation (CPT) module. The TFT module utilizes low-level features from the encoder to refine the details, and the CPT module leverages information from reference images to supplement color information and provide a reference standard for controllable enhancement. By incorporating these modules, we enable a step-by-step refinement process that improves the texture, color, and brightness of the restored image. This design also allows users to adjust the enhancement according to their visual perception, leading to improved customization and user satisfaction.
% flexibly

In summary, CodeEnhance mainly includes two stages. In Stage \uppercase\expandafter{\romannumeral1}, a VQ-GAN \cite{VQGAN} is trained in HQ images by self-reconstruction. In Stage \uppercase\expandafter{\romannumeral2}, we utilize the HQ encoder, the SEM, and the CS to map the LL image to the codebook space. The matched codes are then fed into the frozen HQ decoder and the IFT to generate the enhanced image. 
Our contributions are listed as follows:
% \vspace{-3mm}
\begin{itemize}
    \item We propose CodeEnhance, an innovative LLIE approach that employs a codebook, derived from high-quality images, as prior knowledge. This enables the transformation of low-light images into high-quality ones.

    \item To improve feature matching within the codebook, we introduce the Semantic Embedding Module (SEM) and the Codebook Shift (CS). These components enhance the consistency between the codebook and features learned by the encoder. Additionally, we design an Interactive Feature Transformation (IFT) module to enrich texture information in the decoder.
    
    \item Extensive experiments demonstrate that our proposed method achieves state-of-the-art performance on various benchmarks, including LOL, FiveK, LSRW, LIME, MEF, and NPE.
    % \vspace{-5mm}
\end{itemize}

The remainder of this paper is organized as follows. Section \ref{sec_related_works} reviews the related works. Section \ref{sec_Methodology} presents the model design in detail. Section \ref{sec_experiments} conducts experiments and analyzes the results. Finally, we conclude the paper in Section \ref{sec_conclusion}.

\section{Related Works}\label{sec_related_works}
\subsection{Low-Light Image Enhancement}

Images captured in low-light environments often lack important visual details and exhibit poor image quality \cite{TCI4}\cite{reviewer43}. These types of images can negatively impact the viewer's experience and hinder their comprehension of the image content for further analysis. To address this issue, early researchers explored histogram equalization technology \cite{AHE}, which involves adjusting the illumination and contrast of the image by equalizing pixel intensity. They devised various LLIE methods that focused on the overall image perspective \cite{5773086}, cumulative function \cite{841534}, and penalty terms \cite{4895264}. In addition, many methods based on Retinex theory \cite{Land1971LightnessAR} decompose the image into two components: illumination and image reflection \cite{SSR}.
{With the advancements in deep learning technology \cite{10015857}\cite{reviewer34}\cite{10682473},} LLNet \cite{LLNet} has successfully integrated it into the field of LLIE for the first time utilizing stacked autoencoders. {To further improve image quality, multi-branch \cite{AGLLNet}\cite{reviewer33}, and multi-stage \cite{LRCR} LLIE networks are developed, considering illumination recovery, noise suppression, and color refinement.} SCI \cite{SCI} uses a unique network structure in the training phase, where multiple stages share weights. However, during testing, only one subnetwork is used. SNR \cite{SNR} employs the PSNR distribution map of the image to guide network feature learning and fusion. SMG \cite{SMG} incorporates information about the image structure to enhance the quality of the output image. Moreover, {by combining Retinex theory with deep learning, URetinexNet \cite{URetinex} and DA-DRN \cite{reviewer41} formulated the decomposition problem of Retinex as an implicit prior regularization model.} Retinexformer \cite{Retinexformer} learns the global illumination information by using illumination-guided Transformer \cite{Transformer}. {END \cite{reviewer42} enhances low-light images by dual prior guidance.}
{With the development of diffusion model \cite{reviewer31}\cite{reviewer32}, GSAD \cite{GSAD}, QuadPrior\cite{reviewer44} and JoRes \cite{wu2024jores} leverage the diffusion model to perform LLIE.}

However, these LLIE methods rely on LL image information for enhancement.
This reliance makes them vulnerable to uncertain factors like noise and light sources, causing image artifacts, loss of details, and color distortion.
To address these challenges, we propose a novel model that leverages high-quality prior knowledge to enhance its robustness against such uncertain factors. And a new feature transformation module is introduced to enable the algorithm to better handle variations in the input image.

\subsection{Discrete Codebook Learning} 
Discrete codebook learning was initially introduced in the VQ-VAE \cite{VQVAE}. After that, VQ-GAN \cite{VQGAN} incorporates codebook learning into the GAN framework, enabling the generation of high-quality images. In low-level tasks, codebook learning is employed to mitigate uncertainty during the model learning process by transforming the operational space from the image into compact proxy space \cite{Codeformer}. To improve feature matching, FeMaSR \cite{FeMaSR} introduces residual shortcut connections, RIDCP \cite{RIDCP} proposes a controllable feature matching operation, and CodeFormer \cite{Codeformer} presents a Transformer-based prediction network to obtain the codebook index, respectively. Moreover, LARSR \cite{LARSR} proposes a local autoregressive super-resolution framework based on the learned codebook. VQFR \cite{VQFR} designs parallel decoders to fuse low-level features from the encoder. CodeBGT \cite{ye2024codedbgt} introduces the codebook to improve LLIE model performance.

Building upon the research above, our approach aims to redefine the low-light image enhancement task by introducing the codebook priors and learning the mapping between images and codebook indexes. Additionally, we propose a CS mechanism to fine-tune the original priors. This makes the priors better suitable for different datasets and enables our method to handle various illumination intensities.

\section{Methodology} \label{sec_Methodology}

% The core idea of our CodeEnhance is to improve the mapping from LL images to HQ images by exploiting a discrete representation space while refining the texture and color information for HQ images.
% to redefine the LLIE as a image-to-code model by exploiting a discrete representation space and refine the texture and color information for HQ images.

\begin{figure*}[!t]
    \centering
    \includegraphics[width=1 \textwidth]{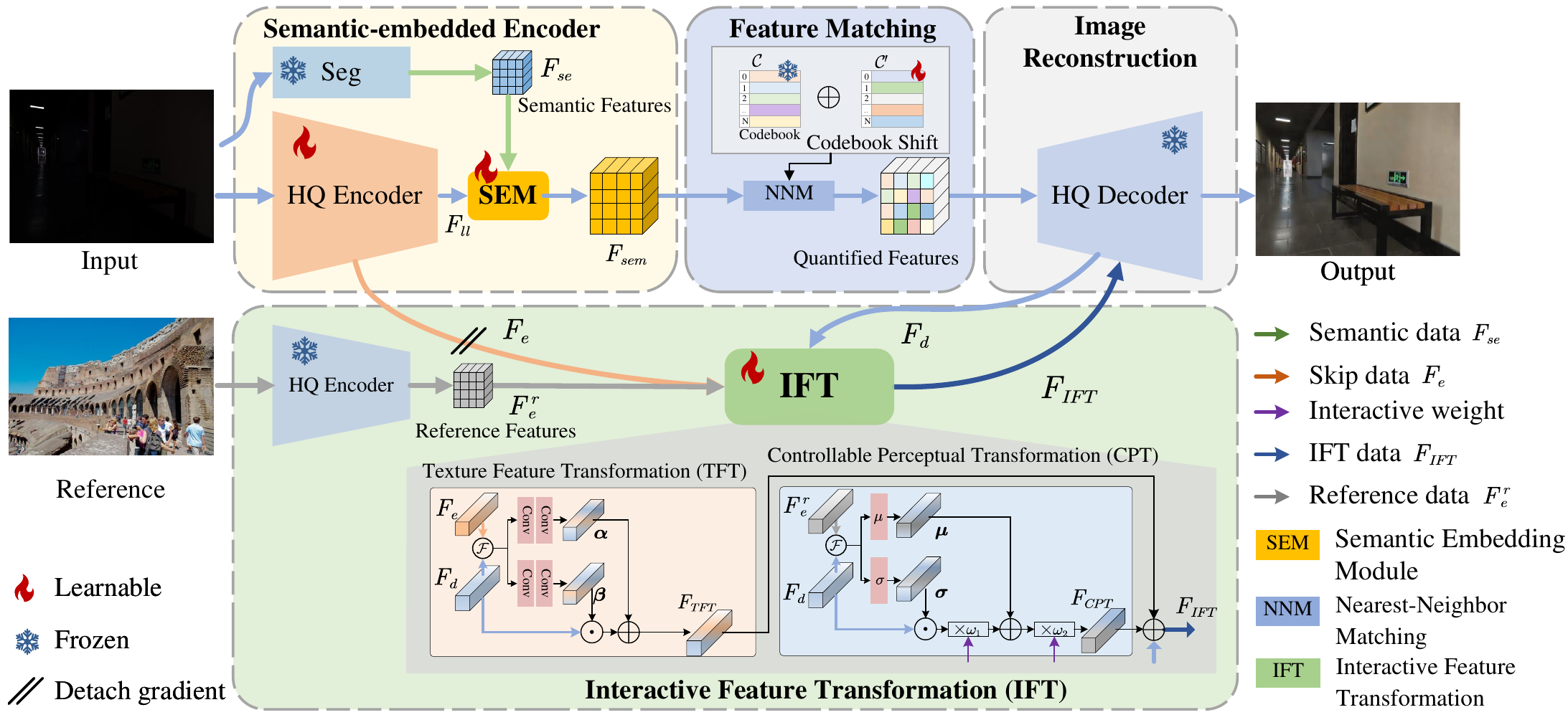}
    \caption{Overview of the proposed CodeEnhance. We utilize the codebook prior and frozen HQ decoder, both of which are learned in Stage \uppercase\expandafter{\romannumeral1}, as a basis for our design. To improve feature learning, an SEM is introduced to bridge the semantic gap between the output of the HQ encoder and the prior. Meanwhile, a CS mechanism is proposed to overcome the distribution shift among datasets and focus more on valuable priors for LLIE.
    Additionally, we propose an IFT module performed by a TFT and a CPT to refine texture, color, and illumination.
}
\label{fig:framework}
\end{figure*}

The core idea of the proposed CodeEnhance is to minimize restoration uncertainty by reframing the LLIE as an image-to-code paradigm. 
The training process for CodeEnhance involves two stages.
In Stage \uppercase\expandafter{\romannumeral1}, a VQ-GAN \cite{VQGAN} is trained to learn a discrete codebook prior and its corresponding decoder. 
As shown in Fig. \ref{fig:framework}, an SEM is first designed to ensure effective feature alignment between the learnable HQ encoder and the codebook in the Stage \uppercase\expandafter{\romannumeral2}.
Then, a CS mechanism is introduced to fine-tune the codebook based on the Stage \uppercase\expandafter{\romannumeral2} dataset to improve feature matching.
Finally, we present an IFT module to improve high-quality details and refine visual perceptual.

\subsection{High-quality Codebook Learning}
We first pre-train a VQ-GAN \cite{VQGAN} using HQ images to learn a discrete codebook. This codebook serves as prior knowledge for the Stage \uppercase\expandafter{\romannumeral2}. The corresponding HQ decoder of the codebook is used to reconstruct HQ images.
Given a HQ image $I_{h}$, it is first processed by the HQ encoder of VQ-GAN to obtain a latent feature $\mathbf{Z}_h\in\mathbb{R}^{m \times n \times d}$. Then, by calculating the distance between each ‘pixel’ $z_h^{(i,j)}$ of $\mathbf{Z}_h$ and the $c_k$ in the learnable codebook $\mathbf{C}=\{c_k \in \mathbb{R}^d\}_{k=0}^{N}$, we replace each $z_h^{(i,j)}$ with the nearest $c_k$ \cite{Codeformer}. After that, the quantized features are obtained $\mathbf{Z}_q \in \mathbb{R}^{m\times n \times d}$:
\begin{equation}\label{qua}
    z^{(i,j)}_{q} = \mathop{\arg \min}\limits_{c_k \in \mathbf{C}} \Vert z^{(i,j)}_h - c_k\Vert_2 ,
\end{equation}
where $N=1024$ is the codebook size, $d=512$ denotes the channel number of $\mathbf{Z}_h$ and $\mathbf{C}$. $m$ and $n$ are the sizes of $\mathbf{Z}_h$ and $\mathbf{Z}_q$. Finally, the reconstructed image $I'_h$ is produced by the HQ decoder. The VQ-GAN is supervised by $\mathcal{L}_{vq}$ \cite{VQGAN}, including L1 loss $\mathcal{L}_{mae}$, codebook matching loss $\mathcal{L}_{cma}$ and adversarial loss $\mathcal{L}_{adv}$:
\begin{equation}\label{vq_loss}
\begin{aligned}
  &\mathcal{L}_{vq} = \mathcal{L}_{mae} + \mathcal{L}_{cma} + \mathcal{L}_{adv},  \\ 
     &\mathcal{L}_{L1} = \Vert I_h - I'_h \Vert_1, \\ 
     &\mathcal{L}_{cma} = \beta \Vert \mathbf{Z}_h - 
     \mathrm{sg}(\mathbf{Z}_q)\Vert_2^2 + \Vert \mathrm{sg}(\mathbf{Z}_h) - \mathbf{Z}_q \Vert_2^2,  \\ 
     &\mathcal{L}_{adv} = \gamma \mathrm{log} \mathcal{D}(I_h) + \mathrm{log}(1 - \mathcal{D}(I'_h)),  
\end{aligned}
\end{equation}
where $\mathcal{D}(\cdot)$ is the discriminator. $\mathrm{sg}(\cdot)$ represents the stop-gradient operator. $\beta = 0.25$ denotes a weight trade-off parameter that governs the update rates of both the encoder and codebook \cite{Codeformer}. $\gamma$ is set to $0.1$ \cite{RIDCP}.

\subsection{Feature Matching via Semantic Embedding and Codebook Shift}
This section focuses on optimizing feature matching in the codebook by designing a semantic-embedded encoder ${\rm SEE}(\cdot)$ comprising an HQ encoder ${\rm E}(\cdot)$ and a key component Semantic Embedding Module (SEM) ${\rm SEM}(\cdot)$. Additionally, we propose a Codebook Shift (CS) mechanism ${\rm CS}(\cdot)$ to ensure distribution consistency between the codebook and the current dataset. These techniques improve accuracy and robustness in feature matching. The Formula \ref{qua} can be redefined as follows:
% The quantized feature $\mathbf{Z}_{ll}^q \in \mathbb{R}^{m\times n \times d}$ of LL image $I_{ll}$ can be obtained by:
\begin{equation}
\begin{aligned}
    & z^{(i,j)}_{q} = \mathop{\arg \min}\limits_{c_k \in \mathbf{C}} \Vert {\rm SEE}(I_{ll})^{(i,j)} - {\rm CS}(c_k)\Vert_2 ,\\
    & {\rm SEE}(I_{ll}) = {\rm SEM}({\rm E}(I_{ll}), {\rm DL}(I_{ll}))
\end{aligned}
\end{equation}
where ${\rm DL}(\cdot)$ represents a pre-trained DeepLab v3+ \cite{deeplab_v3_plus} network, which is used to extract semantic information.

\begin{figure}[!t]
    \centering
    \includegraphics[width=0.48 \textwidth]{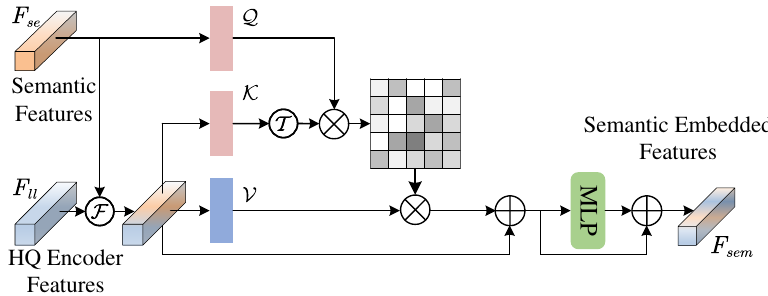}
    \caption{Overview of our Semantic Embedding Module (SEM). }
    \label{fig: sem}
\end{figure}

\noindent\textbf{Semantic Embedding Module.}
Affected by noise and illumination, features in LL images are distorted, resulting in difficult generalization of the original HQ encoder and a semantic gap between the HQ encoder output and the priors. To tackle these challenges, we integrate semantic information and the output of the HQ encoder through a novel SEM. This module compensates for semantic information and enhances the overall learning quality.

As illustrated in Figs. \ref{fig:framework} and \ref{fig: sem}, we use a pre-trained semantic segmentation network (e.g., DeepLab v3+ \cite{deeplab_v3_plus}) to extract the semantic feature $\mathbf{F}_{se}$ from the input image. Moreover, we use the learnable HQ encoder to obtain $\mathbf{F}_{ll}$.
Subsequently, the $\mathbf{F}_{ll}$ and $\mathbf{F}_{se}$ are sent to the SEM.
Within SEM, $\mathbf{F}_{se}$ and $\mathbf{F}_{ll}$ are fused at first, followed by using fused features and $\mathbf{F}_{se}$ to compute an attention-weight map. This map is then applied to the fused features to suppress noise information and obtain semantically embedded features. Lastly, the features are processed by a Multi-Layer Perceptron (MLP) \cite{Transformer}. The SEM can be represented as:
\begin{equation}
    \begin{aligned}
        &\mathbf{M} = \mathcal{Q}(\mathbf{F}_{se})  \mathcal{K}([\mathbf{F}_{se}, \mathbf{F}_{ll}])^T, \\
        &\mathbf{F}' = \mathcal{V}([\mathbf{F}_{se}, \mathbf{F}_{ll}])  \mathbf{M} + [\mathbf{F}_{se}, \mathbf{F}_{ll}], \\
        &\mathbf{F}_{sem} = {\rm MLP}(\mathbf{F}') + \mathbf{F}',
    \end{aligned}
\end{equation}
where $\mathcal{Q}(\cdot)$, $\mathcal{K}(\cdot)$, and $\mathcal{V}(\cdot)$ denote the computing of query, key, and value for obtaining attention-weight map $\mathbf{M}$. $(\cdot)^T$ is transpose operation, and $[\cdot,\cdot]$ represents feature concatenation in channel dimension.

\begin{figure}[!t]
    \centering
    \includegraphics[width=0.48 \textwidth]{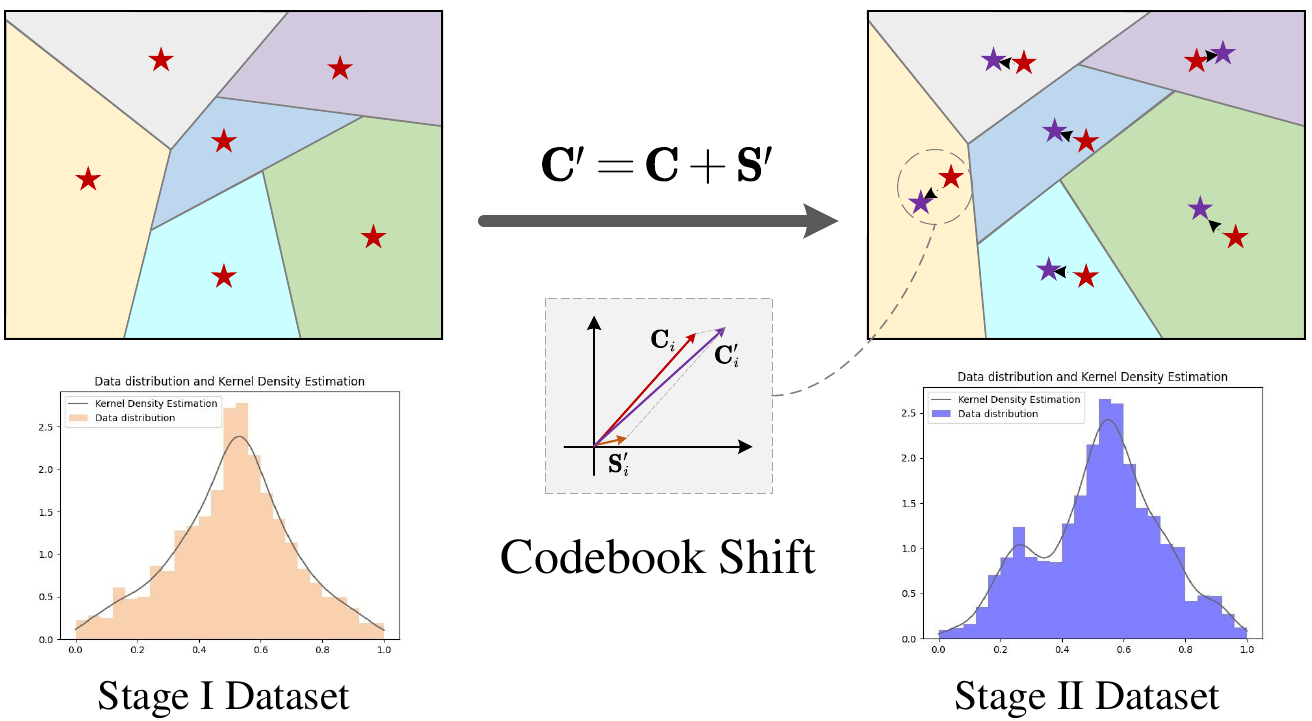}
    \caption{Overview of the proposed Codebook Shift (CS). The Stage \uppercase\expandafter{\romannumeral1} dataset consists of DIV2K \cite{DIV2K} and Flickr2K \cite{Flickr2K}. The Stage \uppercase\expandafter{\romannumeral2} dataset is LSRW Huawei \cite{LSRW}. The static analysis is based on t-SNE and kernel density estimation.}
    \label{fig: cs}
    \vspace{-5mm}
\end{figure}

\noindent\textbf{Codebook Shift.} 
The codebook priors, which consist of HQ image features, act as representative class centers for the Stage \uppercase\expandafter{\romannumeral1} dataset \cite{VQVAE, VQGAN}. 
However, as depicted in Fig. \ref{fig: cs}, there may be a distribution shift between the Stage \uppercase\expandafter{\romannumeral1} and Stage \uppercase\expandafter{\romannumeral2} datasets, presenting a challenge for feature matching. 
Furthermore, different priors have various contributions to feature matching. But most existing methods use the priors without distinction \cite{RIDCP, Codeformer}, which leads to performance degradation in feature matching. 
As shown in Fig. \ref{fig: cs}, to tackle these challenges, we introduce a Codebook Shift (CS) mechanism, which aims to ensure distribution consistency between the codebook and the dataset in Stage \uppercase\expandafter{\romannumeral2} and focus more on valuable priors for LLIE. The CS can be formulated as follows:

\begin{equation}
    \mathbf{C'} = \mathbf{C} + \mathbf{S'},
\end{equation}
where $\mathbf{C'} \in \mathbb{R}^{N\times d}$ denotes the new codebook. $\mathbf{S'} \in \mathbb{R}^{N\times d}$ is the learnable shift.

\subsection{Image Refinement via Interactive Feature Transformation}

We develop Interactive Feature Transformation (IFT) that balances recovering texture details, contrast, and brightness. As shown in Fig. \ref{fig:framework}, the IFT allows us to control the information transmitted to the decoder by Texture Feature Transformation (TFT) and Controllable Perceptual Transformation (CPT). 
The IFT can be formulated as follows:
\begin{equation}
    \mathbf{F}_{IFT} = \mathbf{F}_{TFT} + \mathbf{F}_{CPT},
\end{equation}
where $\mathbf{F}_{TFT}$ and $\mathbf{F}_{CPT}$ are the output of the TFT and CPT, respectively.

\noindent\textbf{Texture Feature Transformation.} The low-level feature $\mathbf{F}_e$ from the HQ encoder contains texture information \cite{CPFE}.
Besides, noise in the low-level feature may affect the quality of reconstruction. 
To address this problem, we design a TFT to effectively incorporate significant texture information and compensate for the loss of details.
As shown in Fig. \ref{fig:framework}, it first fuses $\mathbf{F}_e$ and the decoder features $\mathbf{F}_d$, and then obtains the affine transformation parameters $\bm \alpha$ and $\bm \beta$ to mitigate the impact of noise and extract texture.
Thereby, the TFT can adaptively refine the decoder features by feature affine transformation \cite{SFT} as follows:
\begin{equation}
\begin{aligned}
    \mathbf{F}_{TFT} &= \bm{\alpha} \odot \mathbf{F}_d  + \bm{\beta}, \\
    \bm{\alpha}, \bm{\beta} &= \mathcal{C}([\mathbf{F}_d,\mathbf{F}_e]),
\end{aligned}
\end{equation}
where $\mathcal{C}(\cdot)$ denotes convolution operation, and $\odot$ is element-wise multiplication.

\begin{figure}[!t]
    \centering
    \includegraphics[width=0.48 \textwidth]{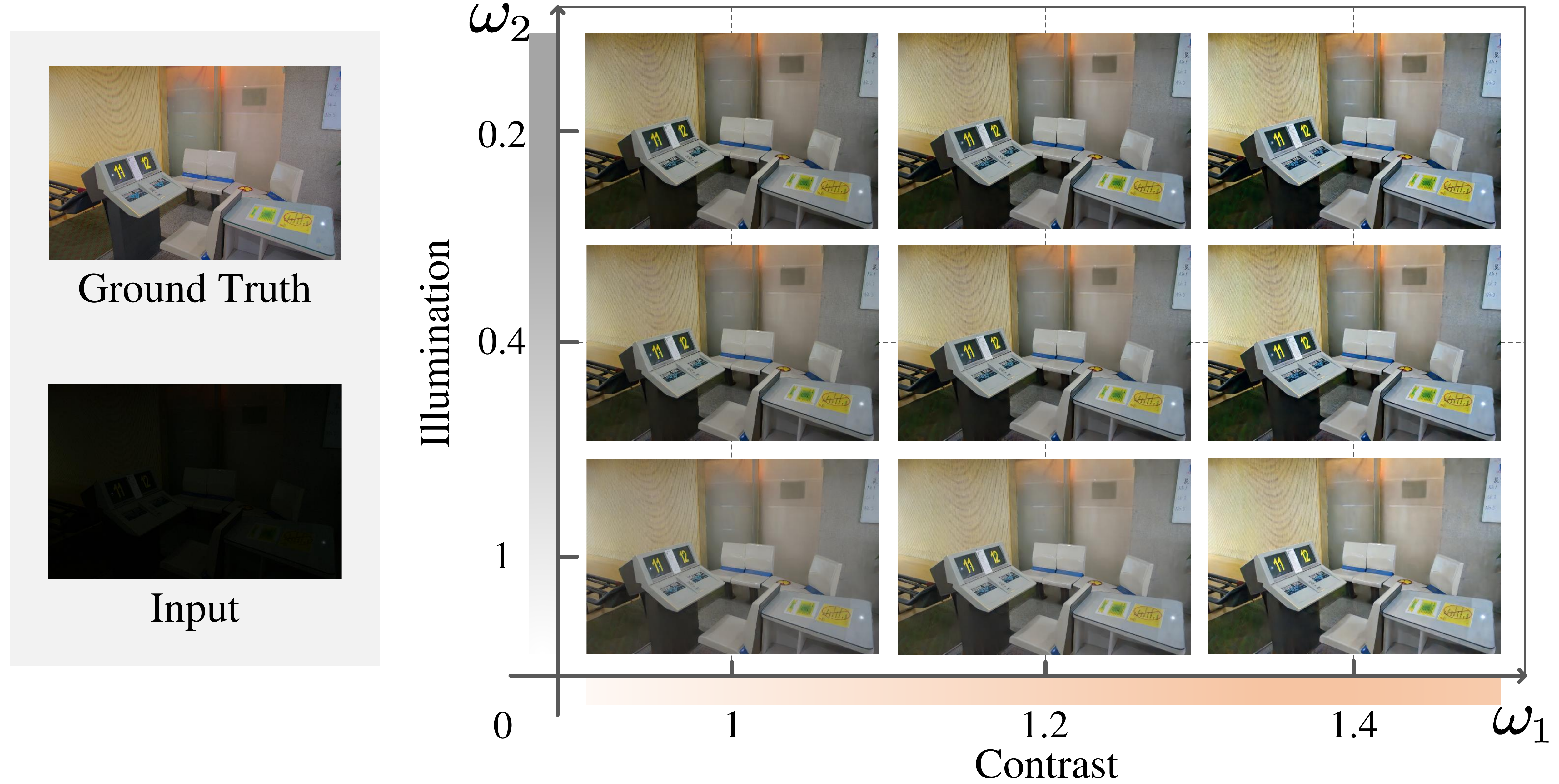}
    % \vspace{-2mm}
    \caption{Results under various adjustment degrees. Our IFT allows users to adjust the enhancement in terms of color and illumination by adjusting the $\omega_1$ and $\omega_2$.}
    \label{fig:IFT}
    \vspace{-4mm}
\end{figure}

\noindent\textbf{Controllable Perceptual Transformation.} 
To enhance the contrast and brightness of the enhanced image, we utilize a high-quality image with proper colors and illumination intensities as a reference for guidance. This reference image serves as valuable information for the model to learn perceptual details. In this way, we introduce a CPT that extracts contrast and brightness information from the reference images and integrates it into the restoration process.
As depicted in Fig. \ref{fig:framework}, the reference image is sent to the pre-trained frozen HQ encoder to obtain reference features $\mathbf{F}_{e}^r$. The CPT can be formulated as follows:
\begin{equation}
\begin{aligned}
    \mathbf{F}_{CPT} &= \omega_1  (\bm{\sigma}  \mathbf{F}_d) + \omega_2  \bm{\mu},
\end{aligned}
\end{equation}
where as shown in Fig. \ref{fig:IFT}, $\omega_1$ and $\omega_2$ are used to control the impact of the CPT in contrast and brightness, respectively. Given a feature $\mathbf{F}'= [\mathbf{F}_d, \mathbf{F}_{e}^{r}]$, $\mathbf{F}' \in \mathbb{R}^{ B \times C \times H \times W}$. The $\bm{\sigma} = [\sigma_{bc}]_{B \times C}$ and $\bm{\mu} = [\mu _{bc}]_{B \times C }$ are computed across spatial locations \cite{AdaIN}, which can be formulated as follows: 
\begin{equation}
\begin{aligned}
    \mathbf{\mu}_{bc} &= \frac{1}{HW}\sum_{h=1}^{H}\sum_{w=1}^{W}\mathbf{F}'_{bchw}, \\
    \mathbf{\sigma}_{bc} &= \sqrt{\frac{1}{HW} \sum_{h=1}^{H}\sum_{w=1}^{W}(\mathbf{F}'_{bchw} - \mu_{bc})^2}.
\end{aligned}
\end{equation}
where Intuitively, a channel feature is regarded as a perceptron that represents a specific style of information \cite{AdaIN}. Images with this style will exhibit higher activation rates on the corresponding channel. By extracting the mean and variance of the channel, the CPT module captures contrast and brightness information associated with the channel. This enables CPT to obtain more nuanced perceptual information while preserving the original content of the image. In essence, CPT controls the transmission of perceptual information in the feature space by utilizing feature statistics (e.g., variance and mean).

\subsection{Training Objectives}
The training objective consists of three loss functions: Feature Matching Loss, Reconstruction Loss, and Adversarial Loss, which can be formulated as follows:
\begin{equation}
    \mathcal{L}_{total} = \mathcal{L}_{fema} + \mathcal{L}_{rec} + \mathcal{L}_{adv} + \lambda_1 \mathcal{L}_{reg},
\end{equation}
where $\mathcal{L}_{adv}$ has been defined in Eq. \ref{vq_loss} and $\mathcal{L}_{reg} = \Vert \mathbf{S}' \Vert_2$ denotes L2 regularization term that controls the influence of $\mathbf{S}'$. $\lambda_1=10^{-4}$ is a hypeparameter. The other two losses are defined as follows.

\noindent\textbf{Feature Matching Loss.} This loss is used to optimize the encoder to learn the mapping between LL images and HQ priors, which can be formulated as follows:
\begin{align}
    \mathcal{L}_{fema} &= \beta \Vert \mathbf{\hat{Z}}_{ll} - 
    \mathrm{sg}(\mathbf{Z}_{gt}^q)\Vert_2^2 + \Vert \phi(\mathbf{\hat{Z}}_{ll}) - \phi(\mathrm{sg}(\mathbf{Z}_{gt})) \Vert_2^2 \notag
\end{align}
where $\phi(\cdot)$ is used to calculate the gram matrix of features. $\mathbf{\hat{Z}}_{ll}$ and $\mathbf{Z}_{gt}^q$ represent the latent features of LL images and quantized features of ground truth, respectively. 

\noindent\textbf{Reconstruction Loss.} This loss focuses on ensuring the enhanced images with a completed structure and impressing visual pleasure, which is performed by L1 loss and perceptual loss: 
\begin{equation}
    \mathcal{L}_{rec} = \Vert I_{gt} - I_{rec} \Vert_1 + \Vert \psi(I_{gt}) - \psi(I_{rec}) \Vert_2^2,
\end{equation}
where $\psi(\cdot)$ indicates the LPIPS function \cite{lpips}.

Overall, the proposed CodeEnhance is constructed through two key modules: feature matching via semantic embedding and codebook shift, and image refinement via interactive feature transformation. This approach redefines LLIE as a task of transforming low-light images into a codebook. Next, we will demonstrate the effectiveness of CodeEnhance through extensive experimental validation.

\section{Experiments}\label{sec_experiments}
\subsection{Implementation Details}
For both the training of VQ-GAN and the proposed method, input image pairs are randomly cropped to obtain the input patches with the size of 256 $\times$ 256. We use ADAM optimizer \cite{Adam} with $\beta_1 = 0.9$, $\beta_2 = 0.999$ and $\varepsilon=10^{-8}$. The learning rate is set to $10^{-4}$. The VQ-GAN is pre-trained on the DIV2K \cite{DIV2K} and Flickr2K \cite{Flickr2K} with 350K iterations. Our CodeEnhance is trained on the low-light datasets with 5K iterations.
All experiments are implemented by the PyTorch framework on an NVIDIA A100 GPU.

\subsection{Datasets and Evaluation Metrics}

\textbf{Low-light Datasets.} We use LSRW \cite{LSRW}, FiveK \cite{FiveK}, LOL \cite{RetinexNet}, and SynLL \cite{AGLLNet} datasets to evaluate the proposed method. 
The LSRW contains Huawei and Nikon datasets. The Huawei consists of training and testing sets with 2450 and 30, respectively. The Nikon includes 3,150 and 20 image pairs for training and testing, respectively.
FiveK \cite{FiveK} dataset consists of 4500 image pairs for training and 500 image pairs for testing.
The LOL dataset includes 485 and 15 image pairs for training and testing, respectively.
The SynLL \cite{AGLLNet} benchmark synthesizes 23,431 short/long exposure image pairs by imposing the degradation of illumination, color, and noise. 22,472 and 959 image pairs are used for training and testing. 
Additionally, we use unpaired datasets including LIME \cite{LIME}, MEF \cite{MEF}, and NPE \cite{NPE} to evaluate generalization performance.

\textbf{Object Detection Dataset.} ExDark \cite{ExDark} is an LL image dataset for object detection. It contains 7,363 images from multi-level illumination and 12 object classes. We partitioned the dataset according to the strategy outlined in \cite{ExDark}, with 3,000 images for training, 1,800 images for validation, and 2563 images for testing.

\textbf{Evaluation Metrics.}
We use the most common Peak Signal-to-Noise Ratio (PSNR), structural similarity \cite{SSIM} (SSIM), Mean Abstract Error (MAE), and Learned Perceptual Image Patch Similarity (LPIPS) \cite{lpips} to measure the quality of the enhanced image. LPIPS considers the human visual system's perception of similarity, which provides a more accurate assessment of the perceived similarity between images. In addition, natural Image Quality Evaluator (NIQE) \cite{NIQE} is used to measure results without ground truth.

\begin{table*}[]
\centering
\caption{Quantitative comparisons on LSRW \cite{LSRW} (Nikon and Huawei), FiveK \cite{FiveK}, LOL \cite{RetinexNet}, and SynLL \cite{AGLLNet}. $\uparrow$ indicates the higher the better, $\downarrow$ indicates the lower the better.}
\resizebox{1 \textwidth}{!}{
\begin{tabular}{c|c|ccccccccccc|
>{\columncolor[HTML]{D9D9D9}}c }
\toprule
Methods                            & Metrics & LIME   & JED    & RetinexNet & KinD   & Zero         & EnGAN  & SNR          & MIRNetv2       & PairLIE      & SMG         & QuadPrior    & Ours            \\
\midrule
                                   & PSNR$\uparrow$    & 14.44  & 14.79  & 13.49      & 15.36  & 15.04        & 14.63  & 16.63        & 17.10           & 15.52        & {\underline 17.26}     & 14.88        & \textbf{17.31}  \\
                                   & SSIM$\uparrow$    & 0.3554 & 0.4600 & 0.2934     & 0.4271 & 0.4198       & 0.3984 & 0.5052       & \textbf{0.5125} & 0.4346       & 0.4927          & 0.4940       & {\underline 0.5061}    \\
                                   & LPIPS$\downarrow$   & 0.3303 & 0.3510 & 0.4041     & 0.3444 & 0.3763       & 0.3248 & 0.5025       & 0.4516          & {\underline 0.3225} & 0.3343          & 0.3648       & \textbf{0.2498} \\ 
\multirow{-4}{*}{Nikon}     & MAE$\downarrow$     & 0.1442 & 0.1572 & 0.1758     & 0.1430 & 0.1506       & 0.1518 & 0.1247       & 0.1233          & 0.1360       & {\underline 0.1134}    & 0.1554       & \textbf{0.1078} \\ \midrule
                                   & PSNR$\uparrow$    & 18.46  & 15.11  & 16.82      & 17.19  & 16.40        & 17.46  & 20.40        & 20.12           & 18.99        & {\underline 20.77}           & 18.35        & \textbf{21.14}  \\
                                   & SSIM$\uparrow$    & 0.4450 & 0.5379 & 0.3951     & 0.4625 & 0.4761       & 0.4982 & 0.6167       & \textbf{0.6317} & 0.5632       & 0.4880          & {\underline 0.6109} & 0.6076          \\
                                   & LPIPS$\downarrow$   & 0.3923 & 0.4327 & 0.4566     & 0.4318 & {\underline 0.3212} & 0.3780 & 0.4879       & 0.4307          & 0.3711       & 0.4193          & 0.4070       & \textbf{0.2840} \\ 
\multirow{-4}{*}{Huawei}    & MAE$\downarrow$     & 0.0950 & 0.1524 & 0.1186     & 0.1155 & 0.1342       & 0.1121 & 0.0784       & 0.0816          & 0.0920       & \textbf{0.0726} & 0.0961       & {\underline 0.0733}    \\\midrule
                                   & PSNR$\uparrow$    & 11.51  & 14.44  & 12.30      & 13.75  & 13.50        & 9.33   & 23.85        & {\underline 24.11}     & 10.55        & 24.08           & 18.11        & \textbf{24.81}  \\
                                   & SSIM$\uparrow$    & 0.6869 & 0.7184 & 0.6874     & 0.7283 & 0.7022       & 0.6459 & {\underline 0.8858} & 0.9007          & 0.6371       & 0.8756          & 0.7806       & \textbf{0.9085} \\
                                   & LPIPS$\downarrow$   & 0.1802 & 0.1947 & 0.2249     & 0.1715 & 0.2084       & 0.2507 & 0.1340       & {\underline 0.0843}    & 0.2695       & 0.1415          & 0.1468       & \textbf{0.0716} \\ 
\multirow{-4}{*}{FiveK}            & MAE$\downarrow$     & 0.2491 & 0.1715 & 0.2026     & 0.1776 & 0.1845       & 0.3254 & 0.0623       & 0.0591          & 0.2710       & {\underline 0.0581}    & 0.1122       & \textbf{0.0531} \\ \midrule
                                   & PSNR$\uparrow$    & 17.18  & 13.69  & 16.77      & 14.78  & 14.86        & 18.68  & {\underline 24.61}        & \textbf{24.74}           & 18.47        & 24.30           & 18.34        & 22.90           \\
                                   & SSIM$\uparrow$    & 0.4747 & 0.6577 & 0.4191     & 0.5520 & 0.5588       & 0.6531 & 0.8419       & \textbf{0.8480}          & 0.7473       & 0.8093          & 0.7785       & {\underline 0.8424}          \\
                                   & LPIPS$\downarrow$   & 0.3419 & 0.2933 & 0.4047     & 0.4506 & 0.3218       & 0.3224 & 0.2064       & {\underline 0.1725}          & 0.2899       & 0.2352          & 0.2358       & \textbf{0.1268}          \\ 
\multirow{-4}{*}{LOL}              & MAE$\downarrow$     & 0.1242 & 0.2108 & 0.1256     & 0.1750 & 0.1846       & 0.1161 & \textbf{0.0552}       & 0.0575          & 0.1153       & {\underline 0.0557}          & 0.1163       & 0.0783          \\ \midrule
                                   & PSNR$\uparrow$    & 13.84  & 13.07  & 12.99      & 16.41  & 13.79        & 16.66  & 22.54        & 22.55           & 16.59        & {\underline 22.94}     & 14.57        & \textbf{23.32}  \\
                                   & SSIM$\uparrow$    & 0.3746 & 0.4017 & 0.3855     & 0.4923 & 0.4217       & 0.5316 & 0.7391       & {\underline 0.7577}    & 0.5706       & 0.7124          & 0.5575       & \textbf{0.7793} \\
                                   & LPIPS$\downarrow$   & 0.4240 & 0.4240 & 0.4568     & 0.4134 & 0.3871       & 0.3979 & 0.3292       & {\underline 0.2424}          & 0.4019       & 0.2827    & 0.3479       & \textbf{0.1904} \\ 
\multirow{-4}{*}{SynLL} & MAE$\downarrow$     & 0.1699 & 0.1963 & 0.1889     & 0.1274 & 0.1881       & 0.1211 & 0.0783       & 0.0627          & 0.1241       & {\underline 0.0603}    & 0.1701       & \textbf{0.0559} \\
\bottomrule
\end{tabular}}
\label{table:compare}
\end{table*}

% \begin{figure}[t]
% \centering
% \subfigure[PSNR $\uparrow$] {\label{fig:heatmap-a}
% \includegraphics[width=2in]{figs/table1_heatmap_PSNR.png}
% }
% % \hspace{-0.1in}
% \subfigure[SSIM $\uparrow$] {\label{fig:heatmap-b}
% \includegraphics[width=2in]{figs/table1_heatmap_SSIM.png}
% }
% % \hspace{-0.1in}
% \subfigure[LPIPS $\downarrow$] {\label{fig:heatmap-c}
% \includegraphics[width=2in]{figs/table1_heatmap_LPIPS.png}
% }
% % \hspace{-0.1in}
% \subfigure[MAE $\downarrow$] {\label{fig:heatmap-d}
% \includegraphics[width=2in]{figs/table1_heatmap_MAE.png}
% }
% \caption{Panels (a)-(d) correspond to Table \ref{table:compare}, which represent comparisons of the PSNR, SSIM, LPIPS, and MAE across different test sets. $\uparrow$ means the larger the better, $\downarrow$ means the smaller the better.}
% \label{fig:heatmap}
% \end{figure}

\begin{figure}[t]
\centering
\subfigure[PSNR $\uparrow$]{
\includegraphics[width=0.22\textwidth]{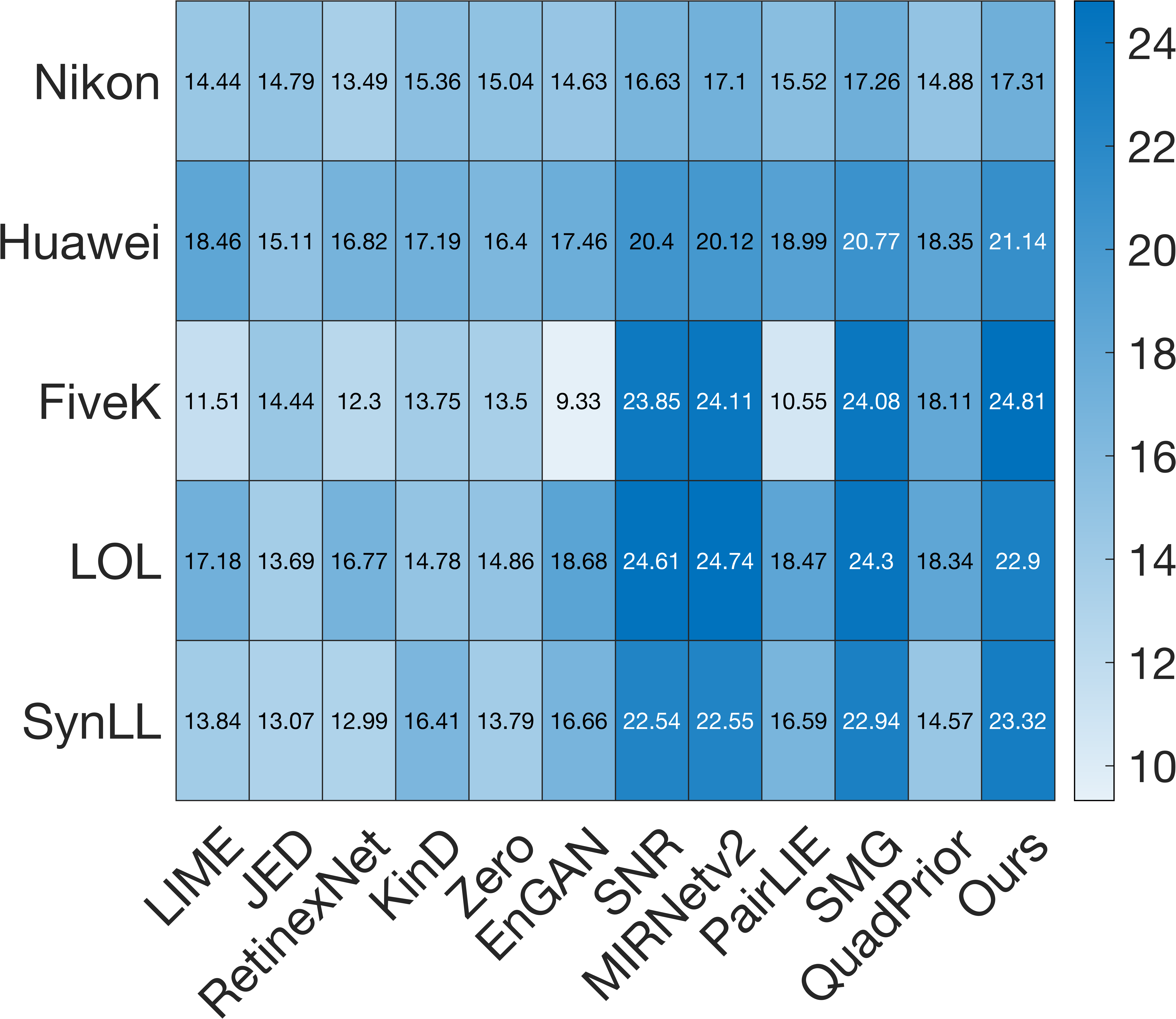}
\label{fig:heatmap-a}
}
\subfigure[SSIM $\uparrow$]{
\includegraphics[width=0.22\textwidth]{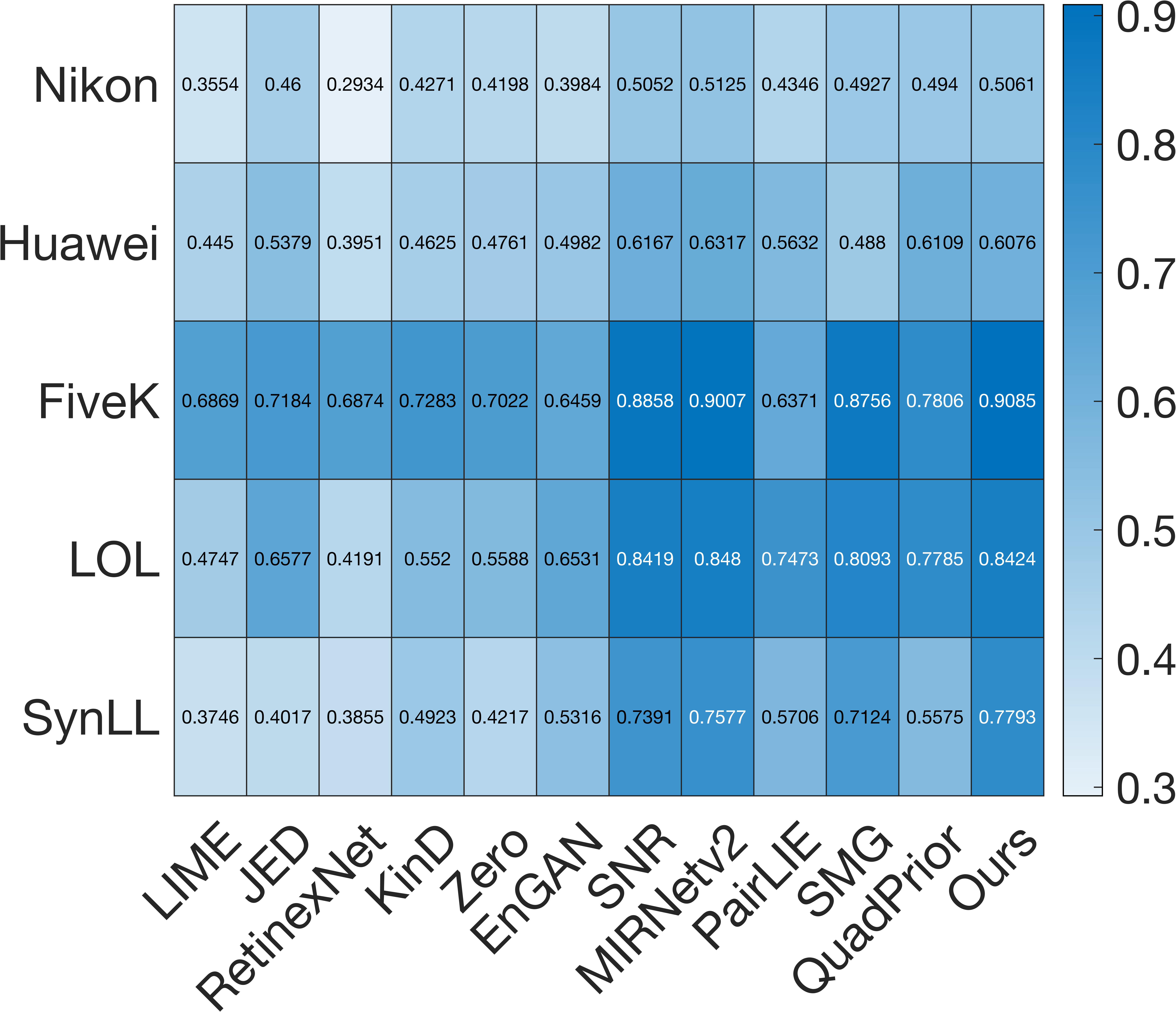}
\label{fig:heatmap-b}
}
\subfigure[LPIPS $\downarrow$]{
\includegraphics[width=0.22\textwidth]{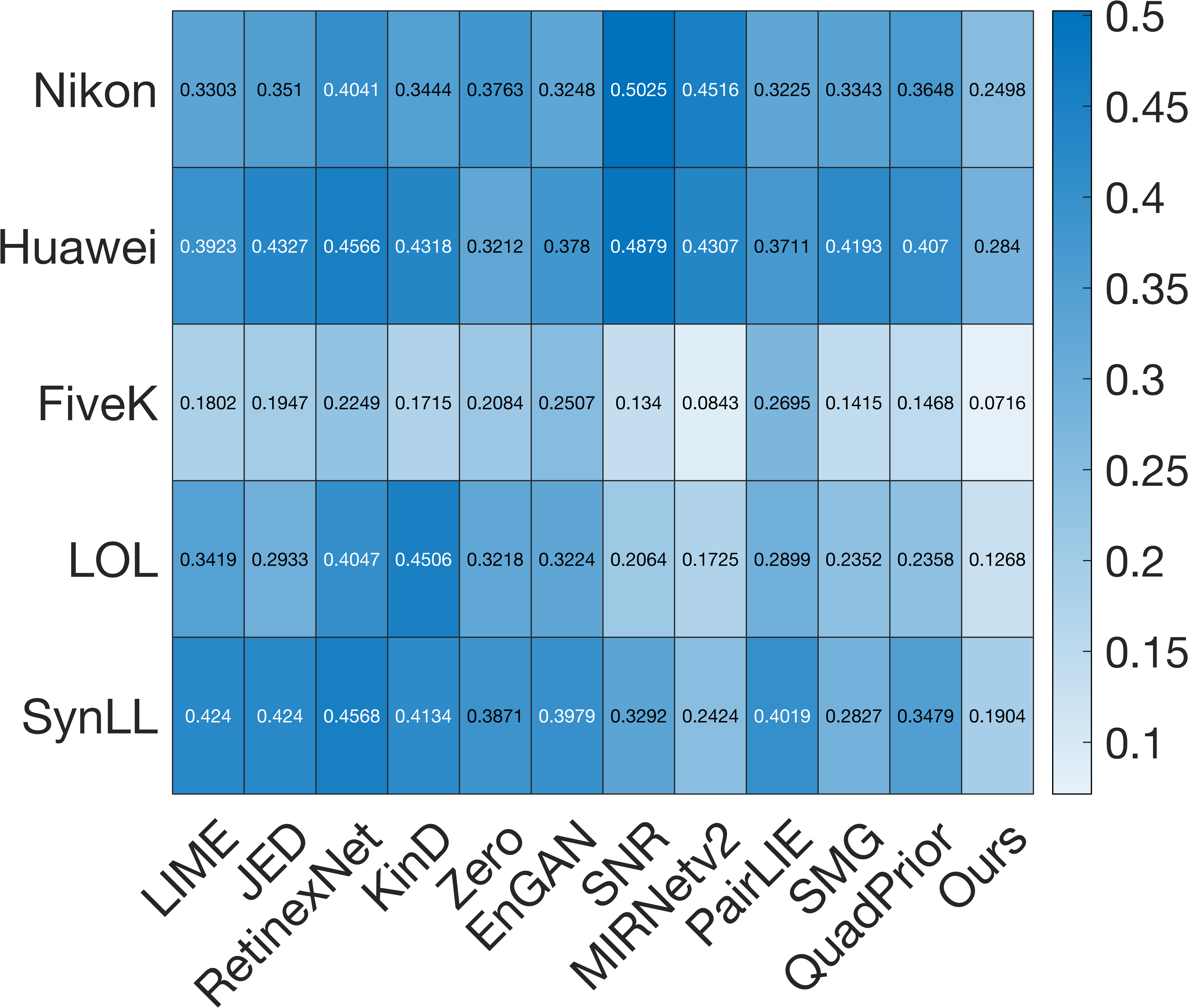}
\label{fig:heatmap-c}
}
\subfigure[MAE $\downarrow$]{
\includegraphics[width=0.22\textwidth]{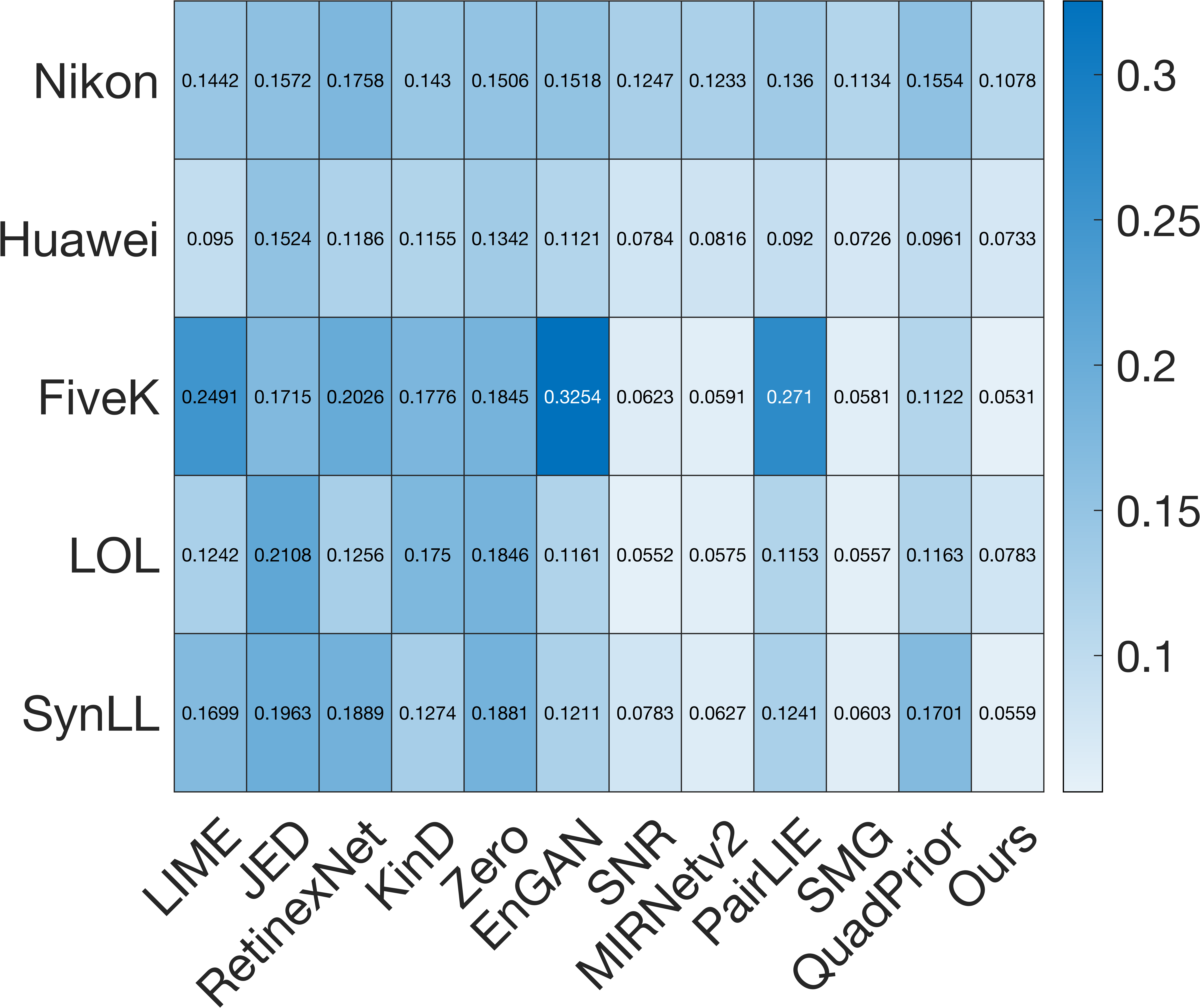}
\label{fig:heatmap-d}
}
\vspace{-2mm}
\caption{Panels (a)-(d) correspond to Table~\ref{table:compare}, showing PSNR, SSIM, LPIPS, and MAE across test sets. $\uparrow$ means higher is better, $\downarrow$ means lower is better.}
\label{fig:heatmap}
\end{figure}

\begin{figure*}[!t]
    \centering
    \includegraphics[width=0.95 \textwidth]{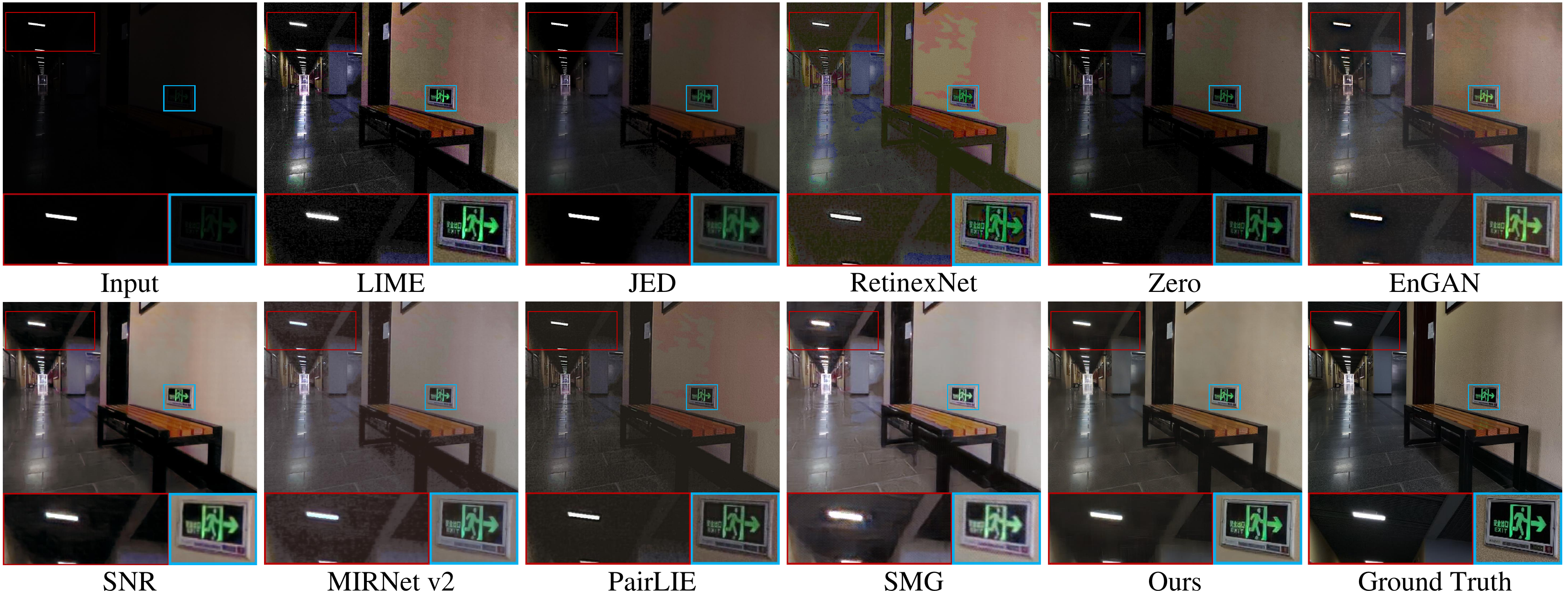}
    \vspace{-2mm}
    \caption{Visual comparisons on the LSRW Huawei \cite{LSRW} dataset. Our results exhibit minimal artifacts and the most accurate color restoration. }
    \label{fig:LSRW_Huawei}
    \vspace{-2mm}
\end{figure*}

\begin{figure*}[!t]
    \centering
    \includegraphics[width=0.95 \textwidth]{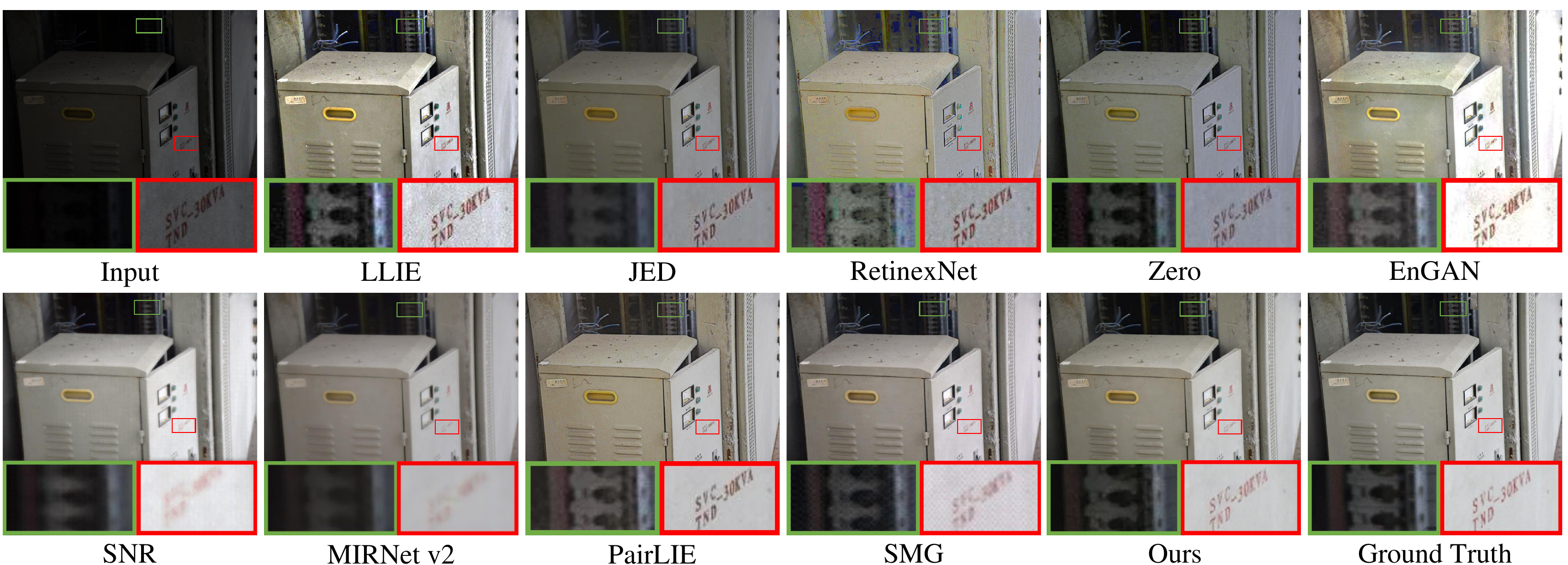}
    \vspace{-2mm}
    \caption{Visual comparisons on the LSRW Nikon \cite{LSRW} dataset. 
    Our results strike a trade-off between noise suppression and texture preservation while maintaining color authenticity.
    } 
    \label{fig:LSRW_Nikon_appendix}
\end{figure*}

\begin{figure*}[!t]
    \centering
    \includegraphics[width=0.95 \textwidth]{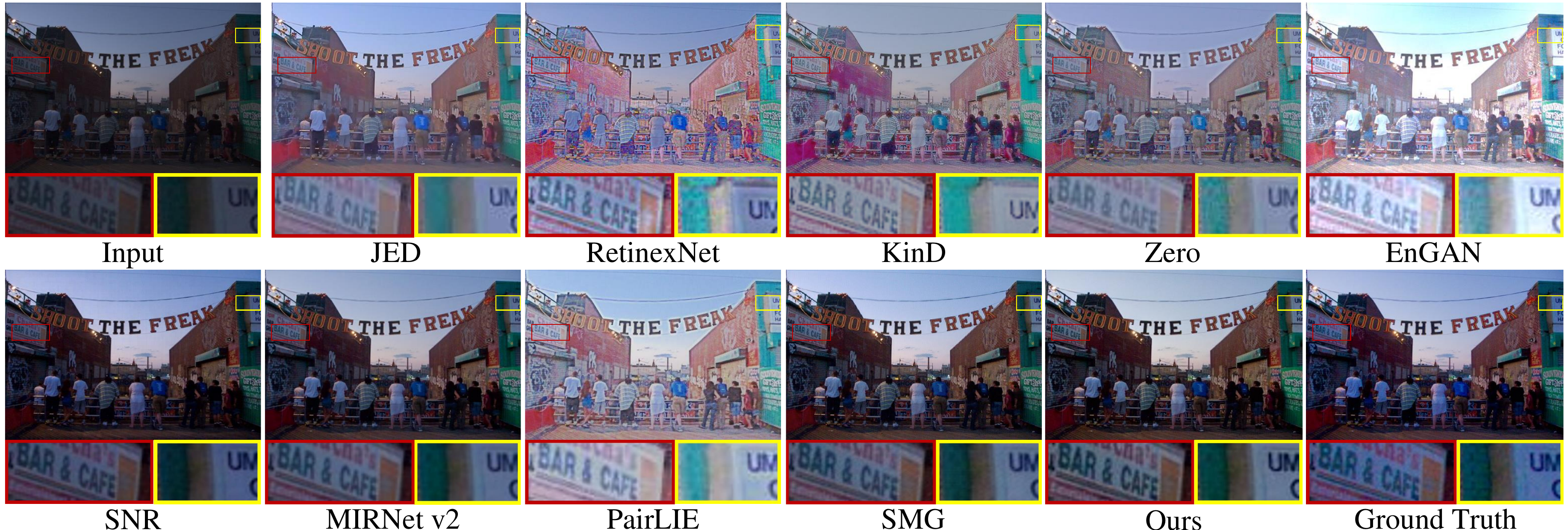}
    \vspace{-2mm}
    \caption{Visual comparisons on the FiveK \cite{FiveK} dataset. Our results are the closest to the ground truth regarding illumination, color, and texture.}
    \label{fig:FiveK}
    % \vspace{-3mm}
\end{figure*}

\begin{figure*}[!t]
    \centering
    \includegraphics[width=0.98 \textwidth]{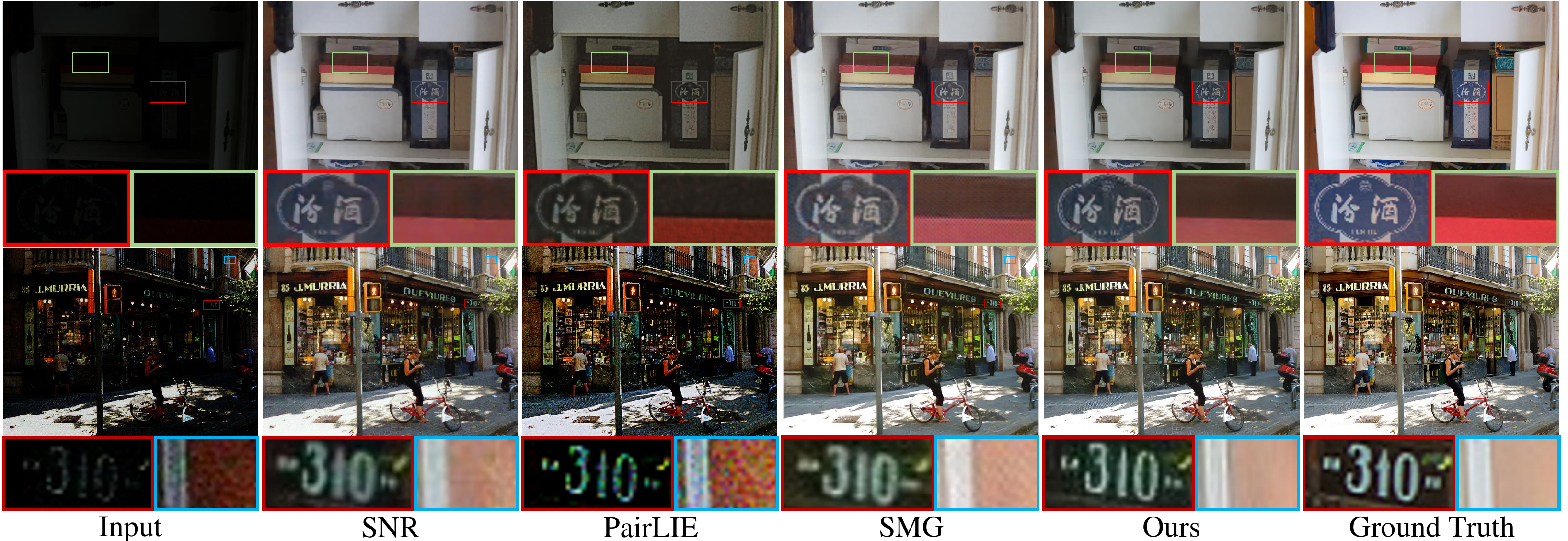}
    \vspace{-2mm}
    \caption{Visual comparisons on the LOL \cite{RetinexNet} and SynLL \cite{AGLLNet}. Our CodeEnhance has less noise and more texture.}
    \label{fig:LOL}
    % \vspace{-2mm}
\end{figure*}

\begin{figure}[t]
\centering
\subfigure[LIME $\downarrow$] {
    \includegraphics[width=0.22\textwidth]{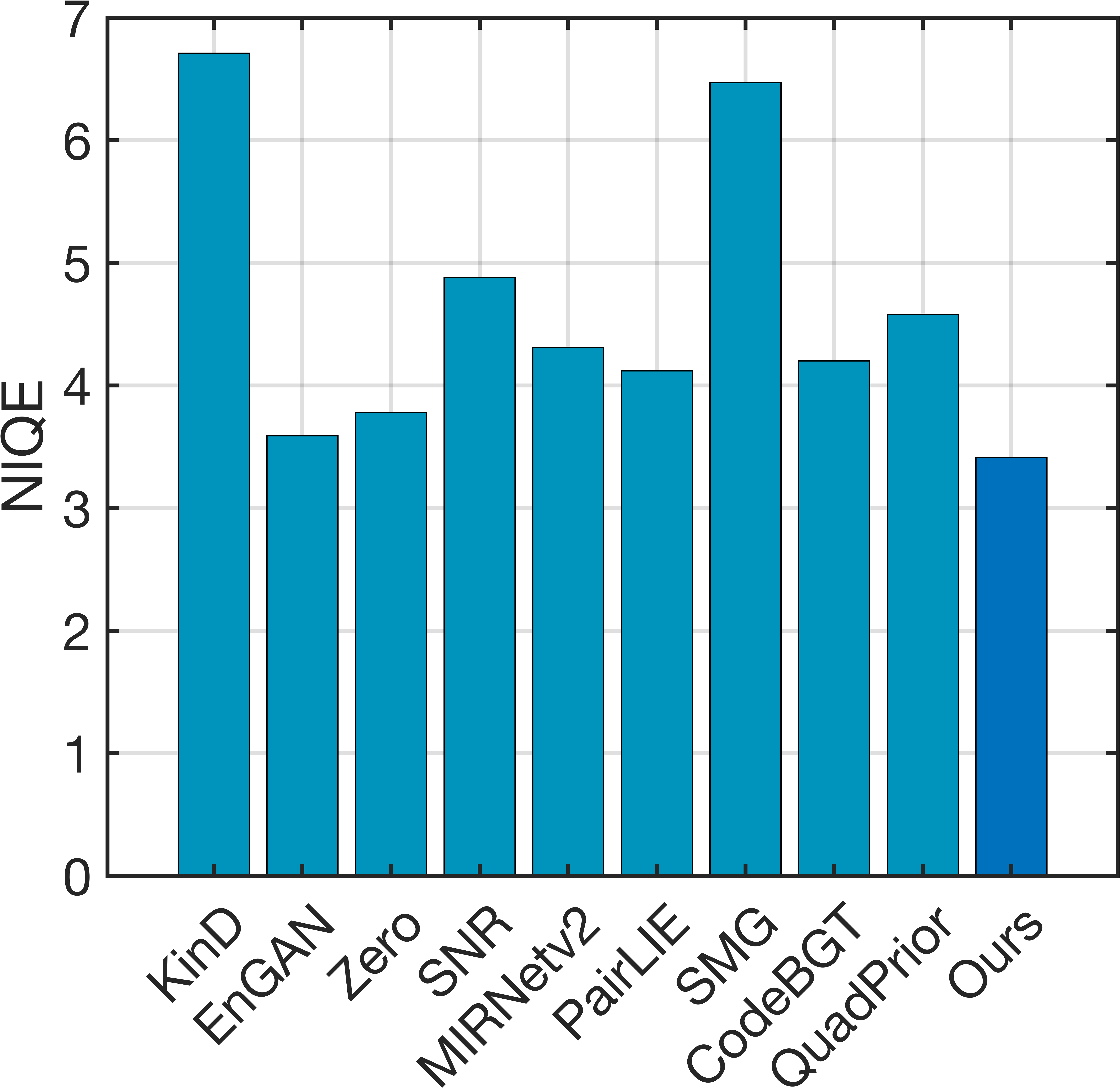}
    \label{fig:bar-a}
}
\subfigure[MEF $\downarrow$] {
    \includegraphics[width=0.22\textwidth]{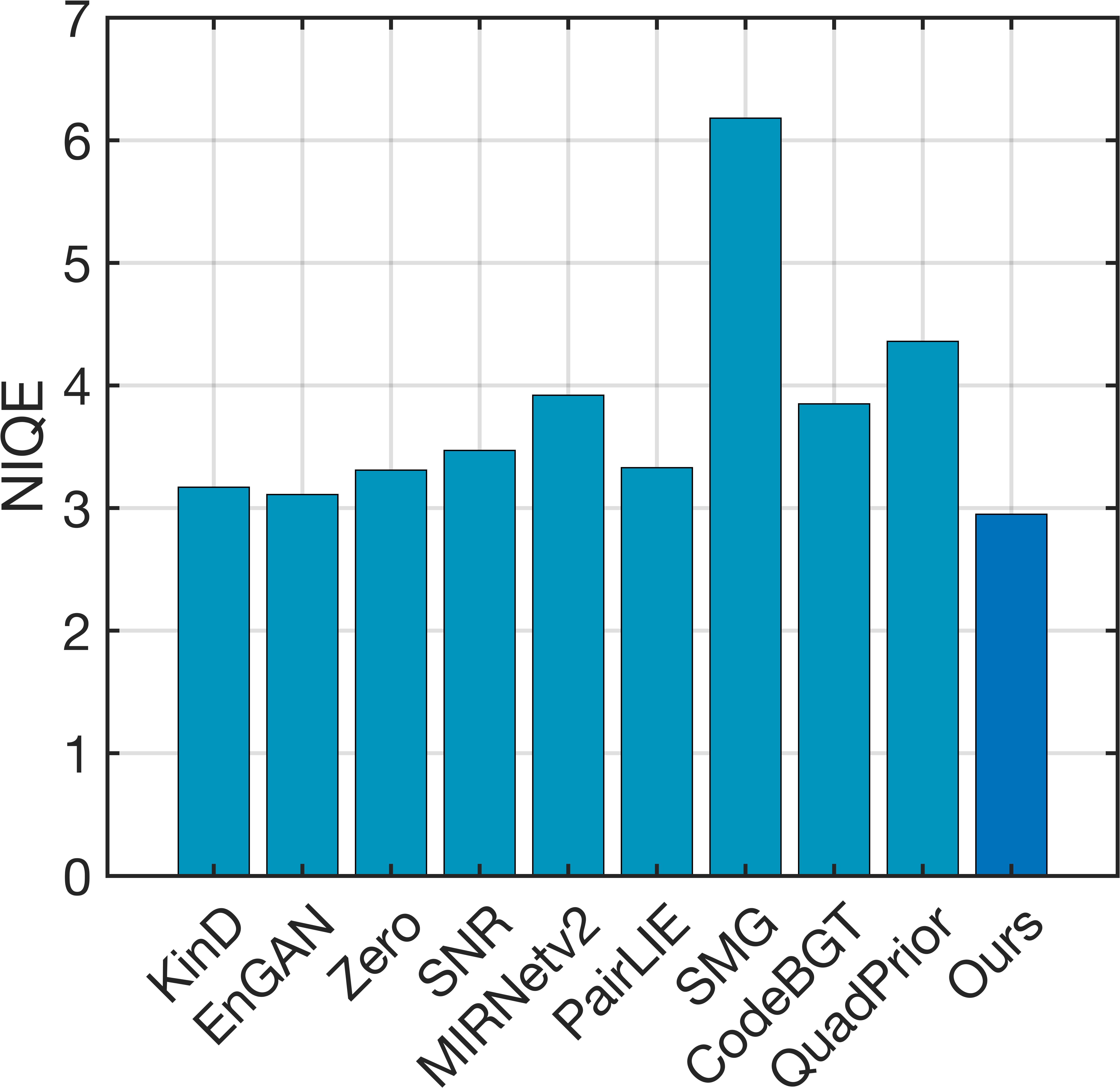}
    \label{fig:bar-b}
}
\subfigure[NPE $\downarrow$] {
    \includegraphics[width=0.22\textwidth]{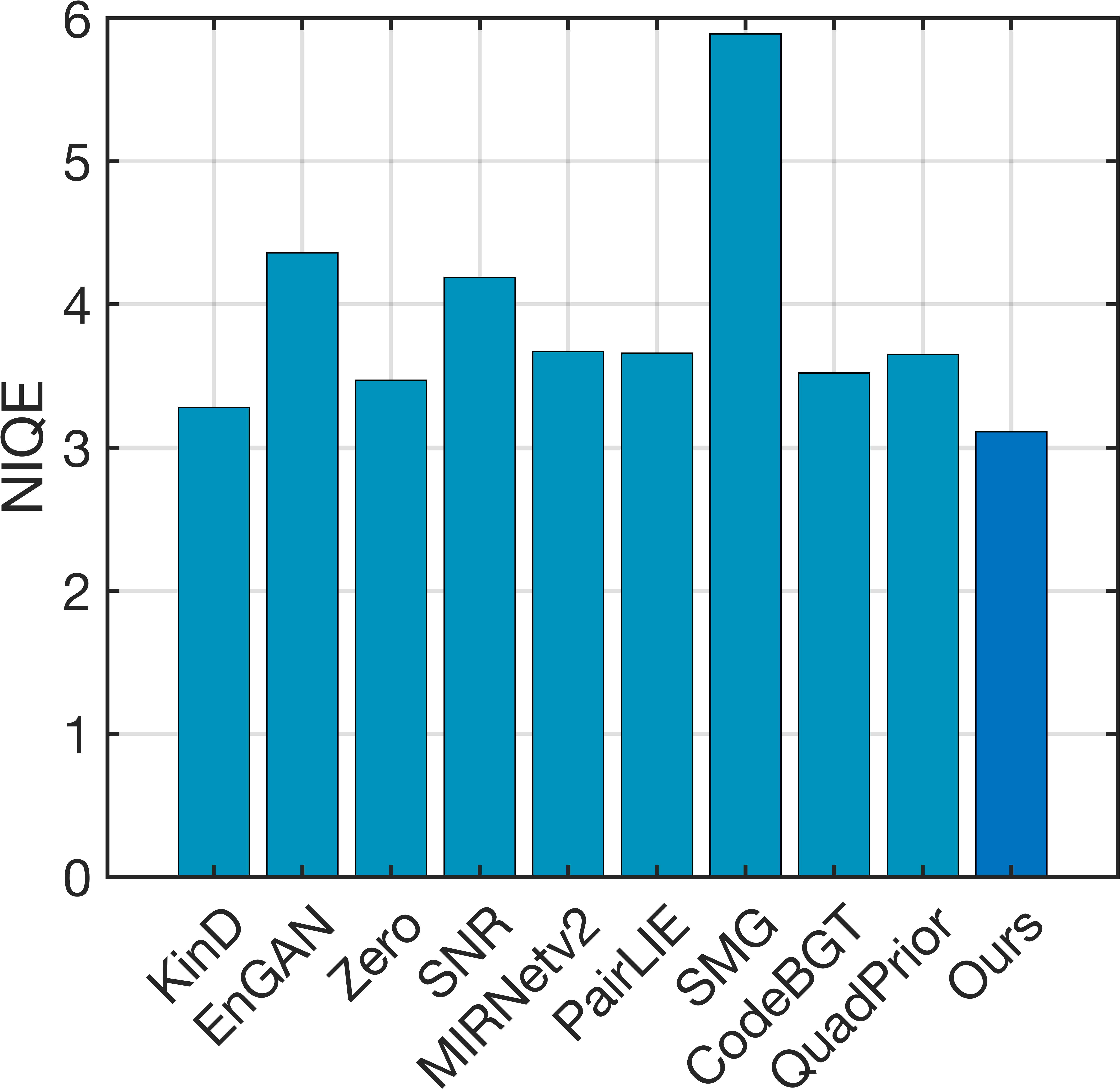}
    \label{fig:bar-c}
}
\subfigure[ExDark $\uparrow$] {
    \includegraphics[width=0.22\textwidth]{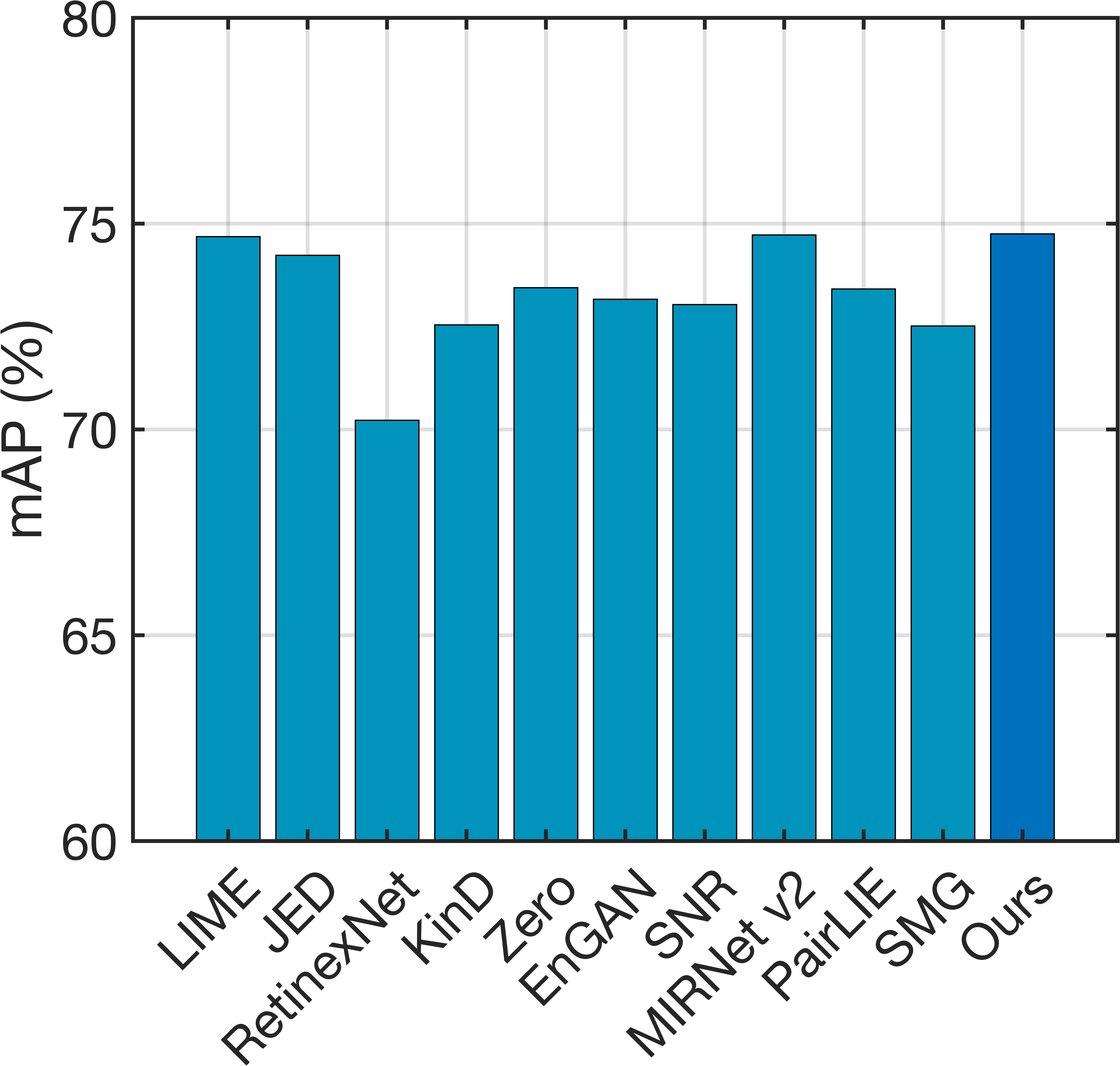}
    \label{fig:bar-d}
}
\vspace{-2mm}
\caption{Panels (a)-(c) correspond to Table~\ref{table:unpaired}, which show NIQE comparisons across LIME~\cite{LIME}, MEF~\cite{MEF}, and NPE~\cite{NPE}. Panel (d) corresponds to Table~\ref{table: ExDark}, showing mAP on ExDark~\cite{ExDark}. $\uparrow$ indicates higher is better, $\downarrow$ indicates lower is better.}
\label{fig:bar}
\end{figure}

\begin{table*}[!t]
\centering
\caption{Comparisons on LIME \cite{LIME}, MEF \cite{MEF}, and NPE \cite{NPE} dataset in terms of NIQE, where the lower the better.}
\resizebox{0.9 \textwidth}{!}{
\begin{tabular}{c|ccccccccc|c}
\toprule
Methods & KinD & EnlightGAN & Zero & SNR  & PairLIE & MIRNetv2 & SMG  & CodeBGT & QuadPrior & \cellcolor[HTML]{D9D9D9}Ours \\
\midrule
LIME    & 6.71 & \underline{3.59}       & 3.78 & 4.88 & 4.31    & 4.12     & 6.47 & 4.20    & 4.58      & \cellcolor[HTML]{D9D9D9}\textbf{3.41} \\
MEF     & 3.17 & \underline{3.11}       & 3.31 & 3.47 & 3.92    & 3.33     & 6.18 & 3.85    & 4.36      & \cellcolor[HTML]{D9D9D9}\textbf{2.95} \\
NPE     & \underline{3.28} & 4.36       & 3.47 & 4.19 & 3.67    & 3.66     & 5.89 & 3.52    & 3.65      & \cellcolor[HTML]{D9D9D9}\textbf{3.11} \\
\bottomrule
\end{tabular}}
\label{table:unpaired}
\vspace{-3mm}
\end{table*}

\subsection{Comparison with State-of-the-Art Methods}
To evaluate the effectiveness of our CodeEnhance, we compare it with several state-of-the-art (SOTA) LLIE methods, including LIME \cite{LIME}, JED \cite{JED}, RetinexNet \cite{RetinexNet}, KinD \cite{KinD}, EnGAN \cite{EnlightGAN}, Zero \cite{Zero}, SNR \cite{SNR}, MIRNet v2 \cite{MIRNetv2}, PairLIE \cite{PairLIE}, SMG \cite{SMG}, from both quantitative and qualitative perspectives.

As shown in Tabel. \ref{table:compare} and Fig. \ref{fig:heatmap}, we compare the proposed CodeEnhance against SOTA LLIE methods regarding PSNR, SSIM, MAE, and LPIPS on the LOL, LSRW (Huawei and Nikon), FiveK, and SynLL datasets. 
{In the case where the LL images contain much noise \cite{AGLLNet}\cite{RetinexNet}, some methods typically over-smooth the noise, leading to higher PSNR \cite{SNR}\cite{SMG}\cite{ MIRNetv2}. However, as shown in Figs. \ref{fig:LSRW_Huawei}-\ref{fig:LOL}, they often lack high-frequency detail information and have poor perceptual quality, while the proposed CodeEnhance can obtain perceptually more convincing results, as indicated by the lowest LPIPS score.
}
In a fully dark scene (Fig. \ref{fig:LSRW_Huawei} and Fig. \ref{fig:LOL}), SNR and SMG improve overall brightness but introduce noise and artifacts, which damage the image quality. In scenes with uneven illumination (Fig. \ref{fig:LSRW_Nikon_appendix} and Fig. \ref{fig:FiveK}), methods like JED, KinD, and PairLIE enhance brightness better but suffer from color distortion and lack of details. SNR and SMG are susceptible to issues such as local color deviation and loss of texture. 
In contrast, our method enhances image illumination, not only suppressing noise and artifacts, but also accurately restoring color and texture information. Intuitively, the text in the figures appears significantly clearer,
significantly improving the visual perception.
This is accomplished by utilizing HQ image priors and incorporating advanced components such as SEM, CS, and IFT in our design.

\begin{figure*}[!t]
    \centering
    \includegraphics[width=0.98 \textwidth]{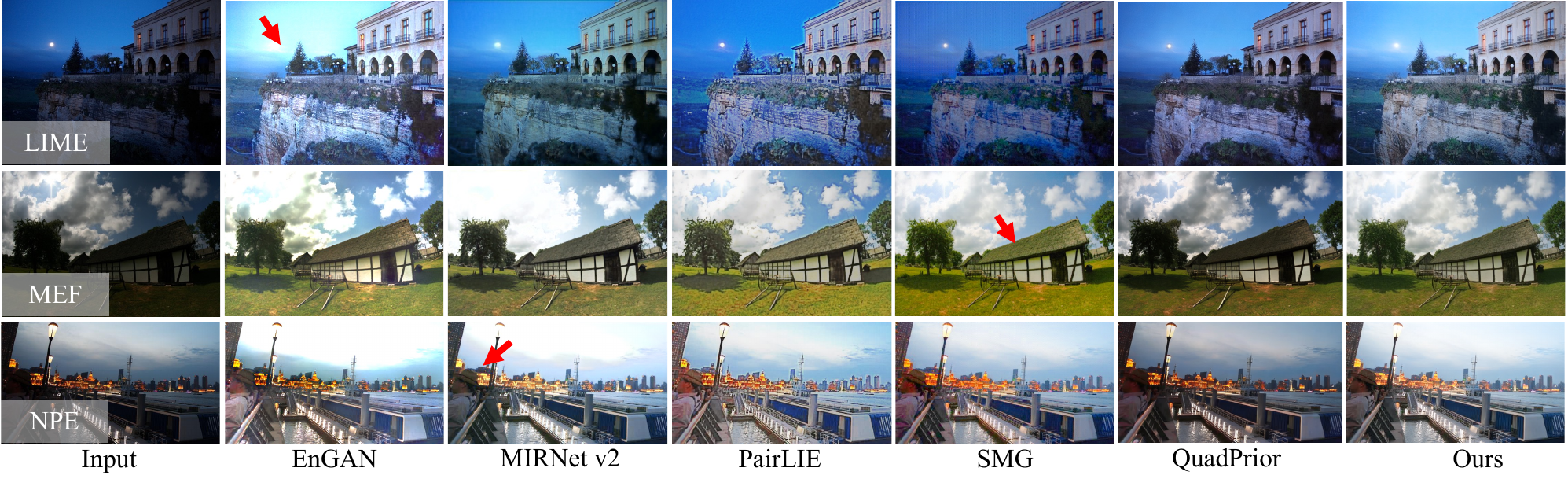}
    \vspace{-2mm}
    \caption{Visual results on the LIME \cite{LIME}, MEF \cite{MEF}, and NPE \cite{NPE} unpaired real-world dataset. Our method enhances the light of input images and maintains natural colors.}
    \label{fig:unpaired}
    \vspace{-3mm}
\end{figure*}

\begin{figure*}[!t]
    \centering
    \includegraphics[width=0.98 \textwidth]{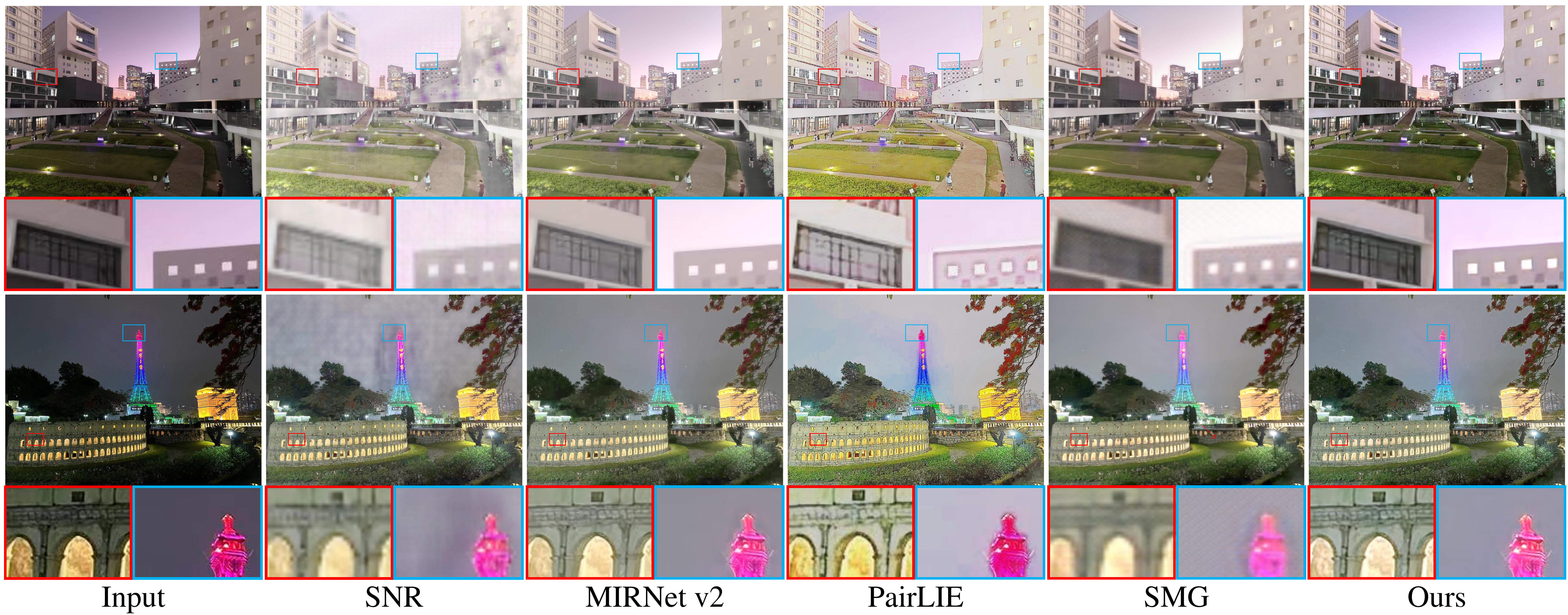}
    \vspace{-2mm}
    \caption{Visual results on the real scene. The results demonstrate the superior performance of our proposed CodeEnhance. Particularly, it excels in preserving texture details and effectively handling gradient change regions.} 
    \vspace{-2mm}
    \label{Real_appendix}
\end{figure*}
\begin{figure*}[!t]
    \centering
    \includegraphics[width=0.85 \textwidth]{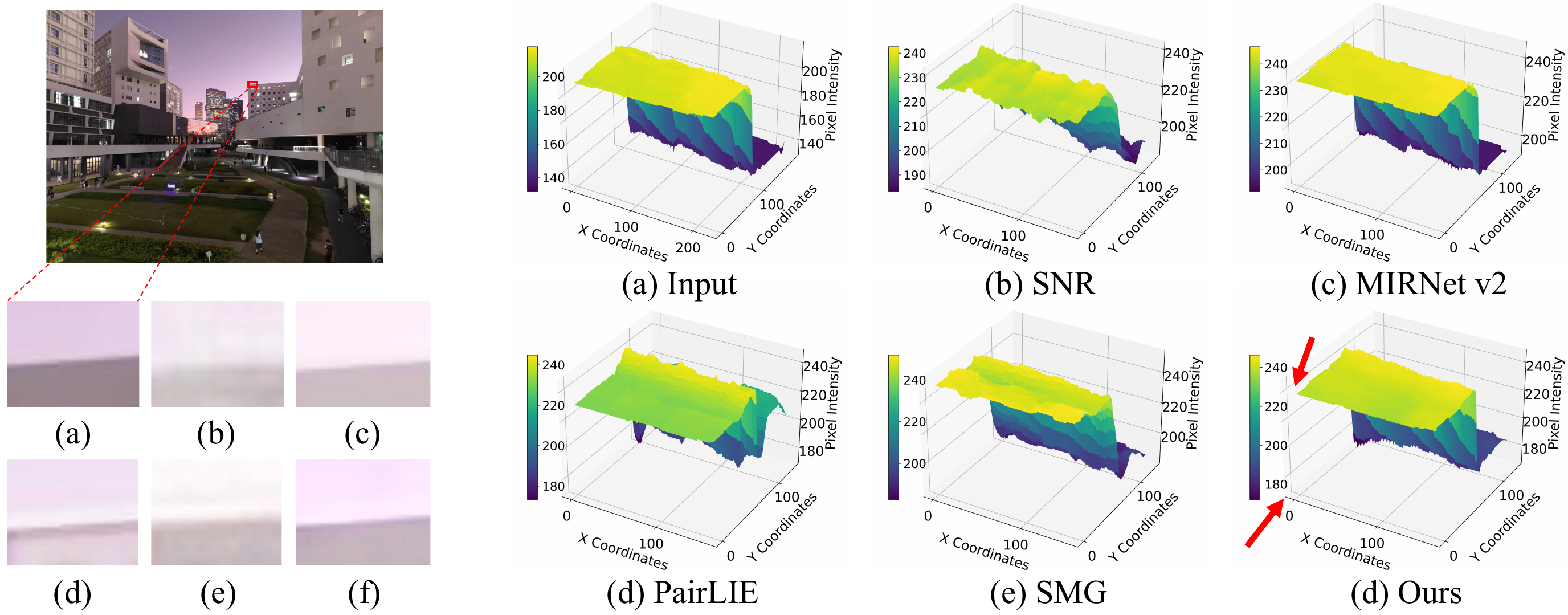}
    \vspace{-2mm}
    \caption{Local first-order differential values correspond to the first row of Fig. \ref{Real_appendix}. The gradients of our results are continuous, broad, and smooth at the edges, indicating superior texture preservation and fewer artifacts} 
    \vspace{-2mm}
    \label{fig:first_order}
\end{figure*}

\begin{table}[!t]
\centering
\caption{Runtime on LSRW Nikon dataset \cite{LSRW}.}
\resizebox{0.48 \textwidth}{!}{
\begin{tabular}{l|ccccc}
\toprule
Methods     & JED   & RetinexNet & KinD   & Zero    & EnGAN     \\ \midrule
Runtime (s) & 6.8399  & 0.8435     & 1.4771   & \textbf{0.0018}  & 0.2151   \\
\midrule
% \hline
Methods     & SNR      & MIRNet v2   & PairLIE   & SMG      & \cellcolor[HTML]{D9D9D9}{Ours}     \\ \midrule
Runtime (s) & 0.3092  & 0.2349     & 0.0525   & 0.5897  & \cellcolor[HTML]{D9D9D9}\underline{0.0330}    \\
\bottomrule
\end{tabular}}
\label{tab_efficient}
\end{table}

To evaluate the generalization performance of the proposed CodeEnhance, we conduct experiments in self-captured night images and unpaired datasets, e.g., LIME \cite{LIME}, MEF \cite{MEF}, and NPE \cite{NPE}. The results are presented in Table \ref{table:unpaired}, Fig. \ref{fig:bar}, Fig. \ref{fig:unpaired}, Fig. \ref{Real_appendix}, and Fig. \ref{fig:first_order}. Compared to other methods, CodeEnhance achieves superior generalization to unseen scenes.
Fig. \ref{fig:unpaired} shows that CodeEnhance effectively avoids artifacts and color deviation, resulting in visually pleasing and artifact-free enhancements. 
Fig. \ref{Real_appendix} and Fig. \ref{fig:first_order} illustrate that our method achieves smooth processing in areas with a wide dynamic range of illumination changes (e.g., gradient sky colors and the transition between buildings and the sky. 
This demonstrates the robustness and effectiveness of our method in handling challenging scenarios with varying illumination conditions.
In addition, Table \ref{tab_efficient} demonstrates that our approach well trade-offs efficiency and image enhancement quality.

{In summary, our improved results stem from the high-quality image priors and three key modules in CodeEnhance. First, the SEM fuses semantic information with low-level features in the encoder, effectively bridging the semantic gap between encoder outputs and the codebook for robust feature matching. Second, the CS mechanism adapts a pre-learned codebook to the unique characteristics of our LLIE dataset, ensuring consistency of distribution and highlighting the most relevant priors. Finally, the IFT module refines texture, color, and brightness by combining low-level encoder features with color guidance from reference images. Together, these design choices reduce uncertainty in the enhancement process and yield high-fidelity, perceptually convincing results.}

\begin{table*}[!t]
\footnotesize
\centering
\caption{The mean Average Precision (mAP) comparisons of low-light object detection on ExDark \cite{ExDark}.}
% \vspace{-2mm}
\resizebox{1 \textwidth}{!}{
\begin{tabular}{l|cccccccccccc|c}
\toprule
Methods    & Bicycle & Boat  & Bottle & Bus   & Car   & Cat   & Chair & Cup   & Dog   & Motor & People & Table & mAP   \\ \midrule
LIME       & 82.75   & 77.59 & 73.92  & 90.12 & 83.6  & 67.22 & 69.51 & 74.01 & 78.29 & 64.05     & 80.95  & 54.24 & 74.68 \\
JED     & 82.74   & 76.74 & 74.32  & 85.57 & 81.07 & 67.89 & 68.56 & 73.64 & 79.79 & 65.84     & 80.4   & 54.21 & 74.23 \\
RetinexNet & 81.97   & 77.59 & 74.08  & 85.68 & 79.61 & 54.92 & 63.32 & 71.28 & 66.39 & 59.08     & 77.59  & 51.17 & 70.22 \\
KinD  & 81.13   & 78.04 & 74.97  & 87.96 & 82.69 & 59.95 & 66.62 & 72.4  & 72.25 & 63.69     & 80.28  & 50.44 & 72.54 \\
Zero    & 83.44   & 77.22 & 75.17  & 89.7  & 82.4  & 62.67 & 66.8  & 73.17 & 75.68 & 64.16     & 80.68  & 50.22 & 73.44 \\
EnGAN  & 81.68   & 76.39 & 74.83  & 89.44 & 80.16 & 63.81 & 65.57 & 72.53 & 77.18 & 64.24     & 80.55  & 51.51 & 73.16 \\
SNR  & 83.65   & 77.44 & 72.97  & 89.01 & 82.7  & 62.64 & 66.51 & 71.02 & 74.91 & 64.74     & 80.19  & 50.55 & 73.03 \\
MIRNet v2 & 85.02   & 79.35 & 74.27  & 90.68 & 82.91 & 66.36 & 67.87 & 70.14 & 79    & 64.76     & 80.67  & 55.6  & \underline{74.72} \\
PairLIE & 82.87   & 77.38                    & 72                         & 88.11                   & 82.38                   & 62.03                   & 70.23                     & 70.44                   & 76.31                   & 64.1                          & 80.11                      & 54.95                     & 73.41                   \\
SMG   & 82.05   & 76.51 & 71.71  & 88.07 & 79.11 & 64.97 & 66.84 & 68.32 & 77.69 & 63.16     & 77.88  & 53.78 & 72.51 \\
\rowcolor[HTML]{D9D9D9} 
Ours & 84.08 &	78.58&	75.08&	88.95&	82.33&	67.12&	68.62&	74.19&	78.85&	63.63&	81.71&	53.83&	\textbf{74.75} \\
\bottomrule
\end{tabular}}
% \vspace{-4mm}
% \label{table:compare}
% \vspace{-2mm}
\label{table: ExDark}
\end{table*}

\begin{figure*}[!t]
    \centering
    \includegraphics[width=0.98 \textwidth]{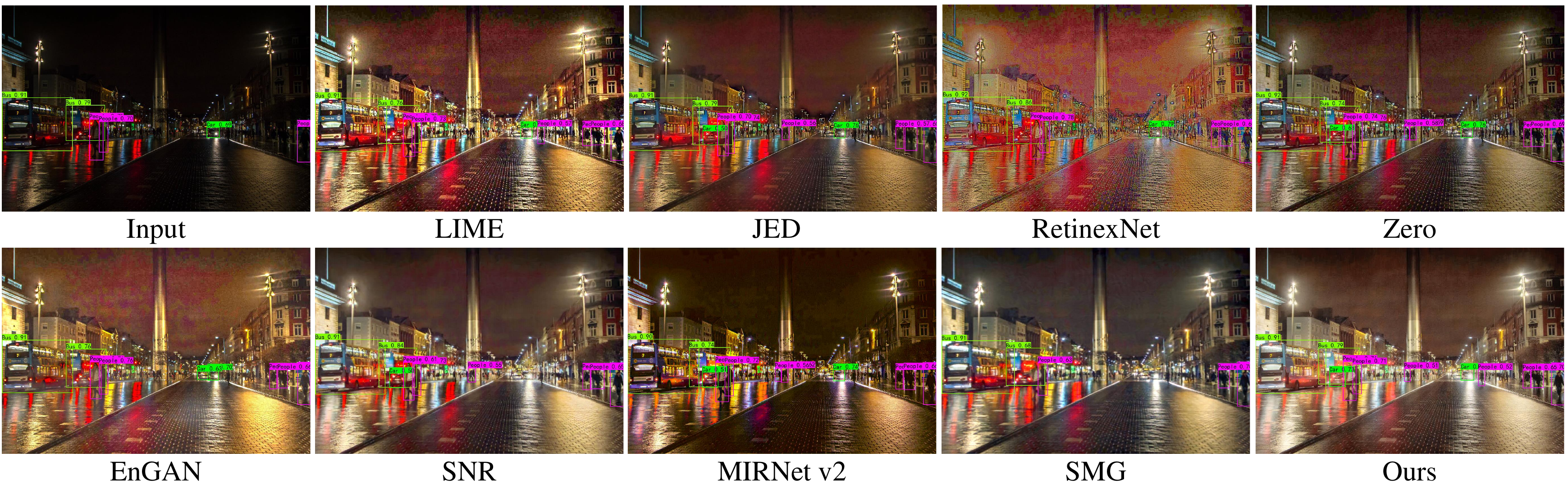}
    \vspace{-2mm}
    \caption{Comparisons of the improvement on object detection tasks \cite{ExDark}. Our method not only enhances the image quality but also improves the performance of object detection algorithms.}
    \label{fig:ExDark}
    \vspace{-4mm}
\end{figure*}

\subsection{Low-light Object Detection}
Although LLIE preprocessing improves the overall illumination and visual perception of low-light images, it may also potentially degrade image features. Specifically, when the features of enhanced images are not effectively preserved or enhanced, the LLIE methods negatively impact the performance of high-level tasks. To mitigate this impact, it is important to consider enhancing edges, textures, and object details \cite{AdaYOLO}.

To assess the impact of our method in enhancing the performance of high-level tasks under low-light conditions, we utilize YOLOv7 \cite{YOLOv7} as the object detector and train it on the ExDark dataset, which was preprocessed using the LLIE methods. The comparison of Average Precision (AP) is presented in Table \ref{table: ExDark} and Fig. \ref{fig:bar-d}, revealing that our method yields images that better facilitate the learning of the target detection algorithm. Furthermore, Fig. \ref{fig:ExDark} demonstrates the effectiveness of our method in suppressing noise while increasing overall brightness in images with complex scenes. As a result, high-quality enhanced images obtained by the proposed CodeEnhance enable YOLOv7 to extract more valuable information and achieve better detection performance.

\subsection{Ablation Study}
To verify the effectiveness of the core modules of our CodeEnhance, we conduct a series of ablation studies. The Baseline model is constructed by a VQ-GAN \cite{VQGAN}.

\textbf{Study of SEM.}
According to the Table \ref{tab: sturcture ablation}, we first assess the proposed SEM. The results show that SEM effectively enhances the model's performance. For example, when compared to Exp. (a) using the baseline model, PSNR and SSIM increased by 0.33 and 0.0152, respectively. These results indicate that SEM can acquire high-quality image features, thereby enhancing feature matching accuracy to improve overall model performance.

\begin{figure}[!t]
    \centering
    \includegraphics[width=0.48 \textwidth]{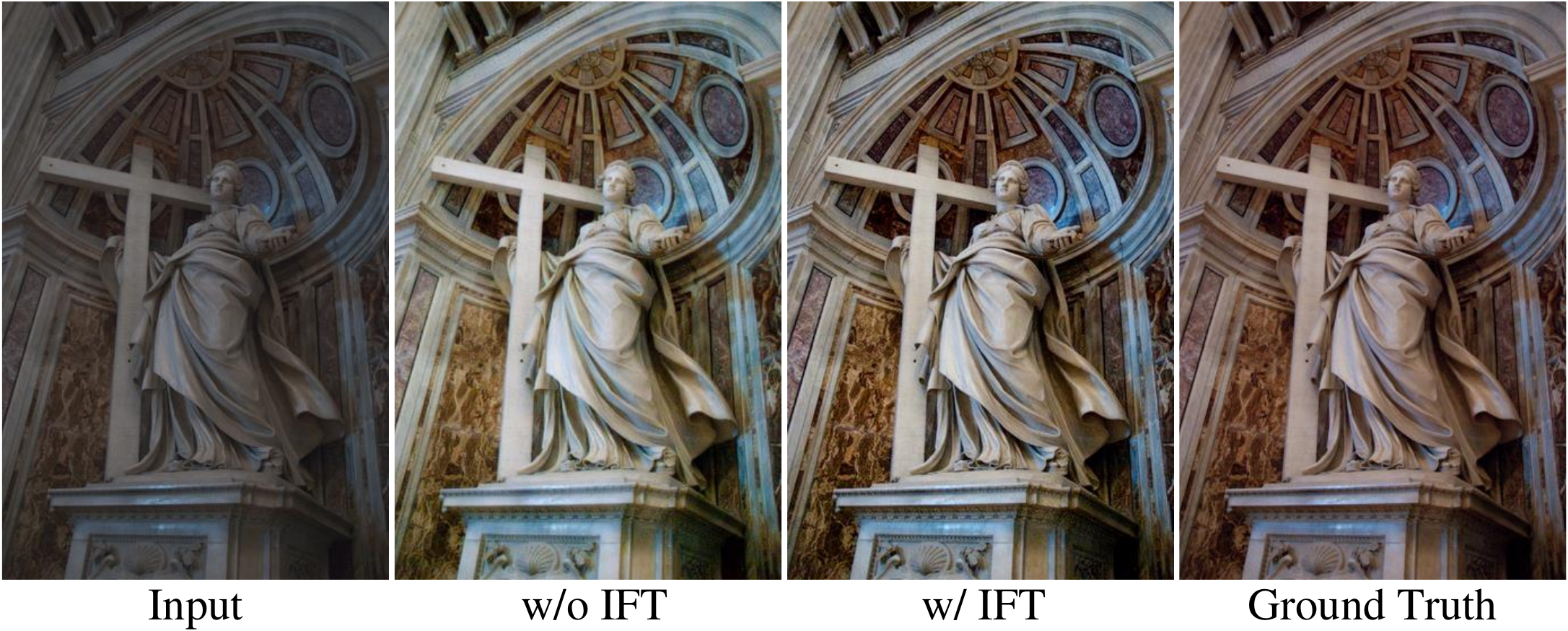}
    \caption{Visualized ablation of IFT (top) on FiveK \cite{FiveK}. and corresponding visual of feature maps (bottom). The higher the feature activation, the greater the feature score. It is evident that our IFT enhances CodeEnhance's ability to capture more texture information.
     }
    \label{fig:IFT_ablation_img}
\end{figure}

\textbf{Study of IFT.}
IFT consists of a TFT and a CPT, and we will individually evaluate their effectiveness. 
Figs. \ref{fig:IFT} and \ref{fig:IFT_ablation_img}, Table \ref{tab: sturcture ablation} and Table \ref{tab: sturcture ablation2} demonstrate that the proposed IFT and CPT significantly improve the performance of our CodeEnhance.
For instance, when compared to Exp. (b) in Table \ref{tab: sturcture ablation}, there is an increase in PSNR of 1.36 and an increase in SSIM of 0.0167. These findings validate the effectiveness of our IFT in effectively controlling texture, color, and illumination information, thereby enhancing the overall model performance.
% \ref{fig:IFT_ablation}

\begin{figure*}[!t]
    \centering
    \includegraphics[width=0.8 \textwidth]{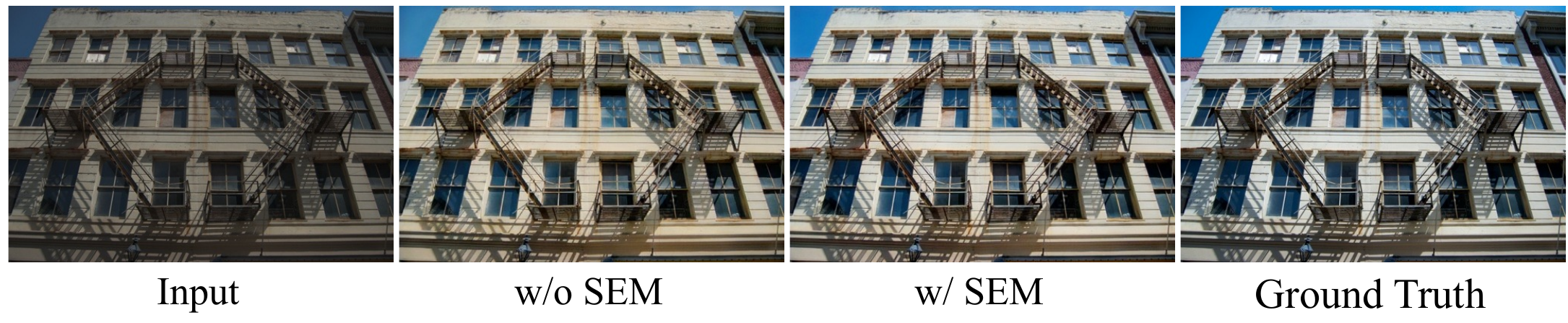}
    \caption{{Visualized ablation of SEM on FiveK \cite{FiveK}. }}
    \label{fig:SEM_ablation_img}
\end{figure*}

\begin{figure}[!t]
    \centering
    \includegraphics[width=0.48 \textwidth]{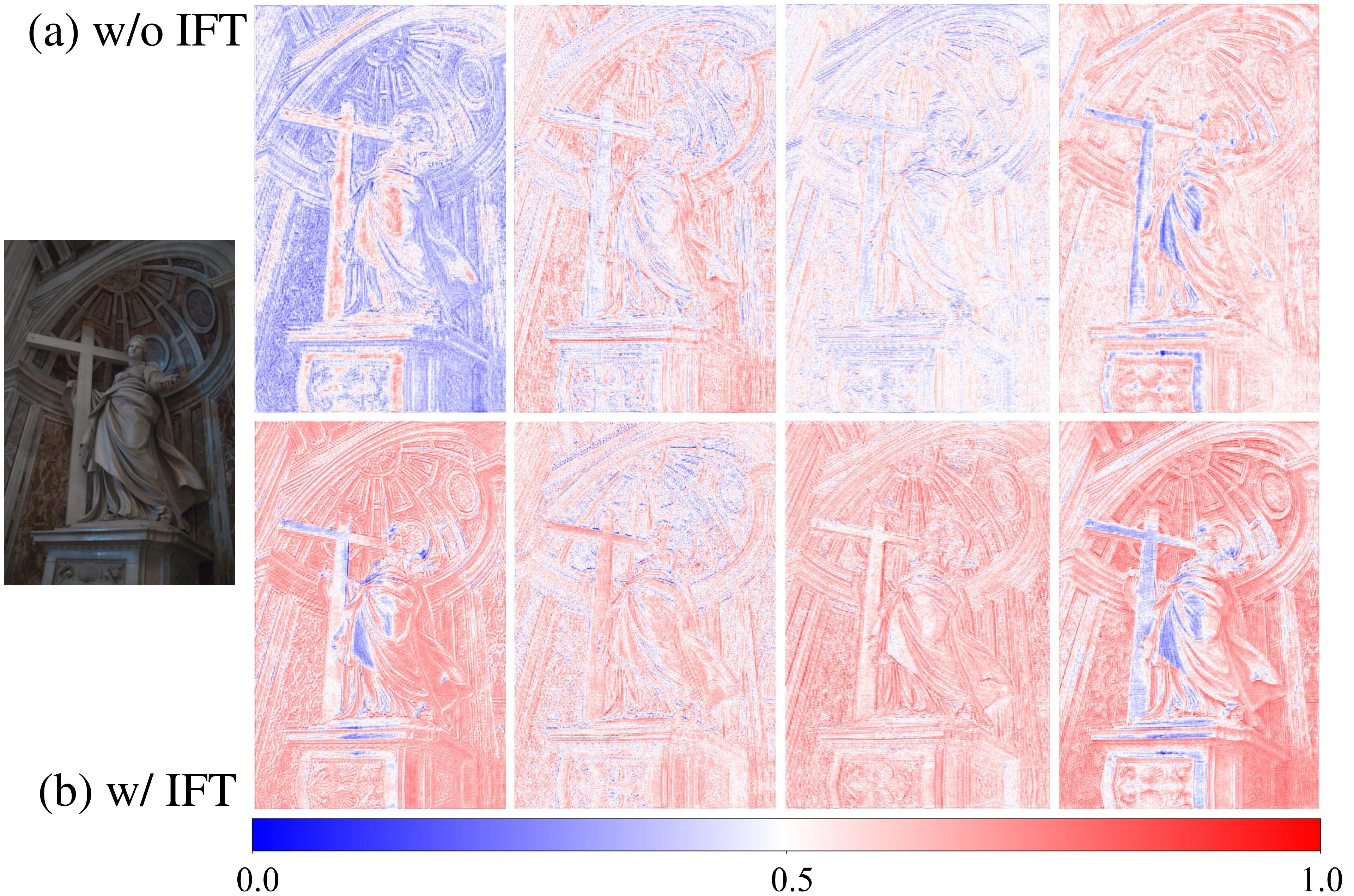}
    \caption{Visual of feature maps. The higher the feature activation, the greater the feature score. It is evident that our IFT (bottom) enhances CodeEnhance's ability to capture more texture information.}
    \label{fig:IFT_ablation}
    \vspace{-4mm}
\end{figure}

\begin{figure}[!t]
    \centering
    \includegraphics[width=0.45 \textwidth]{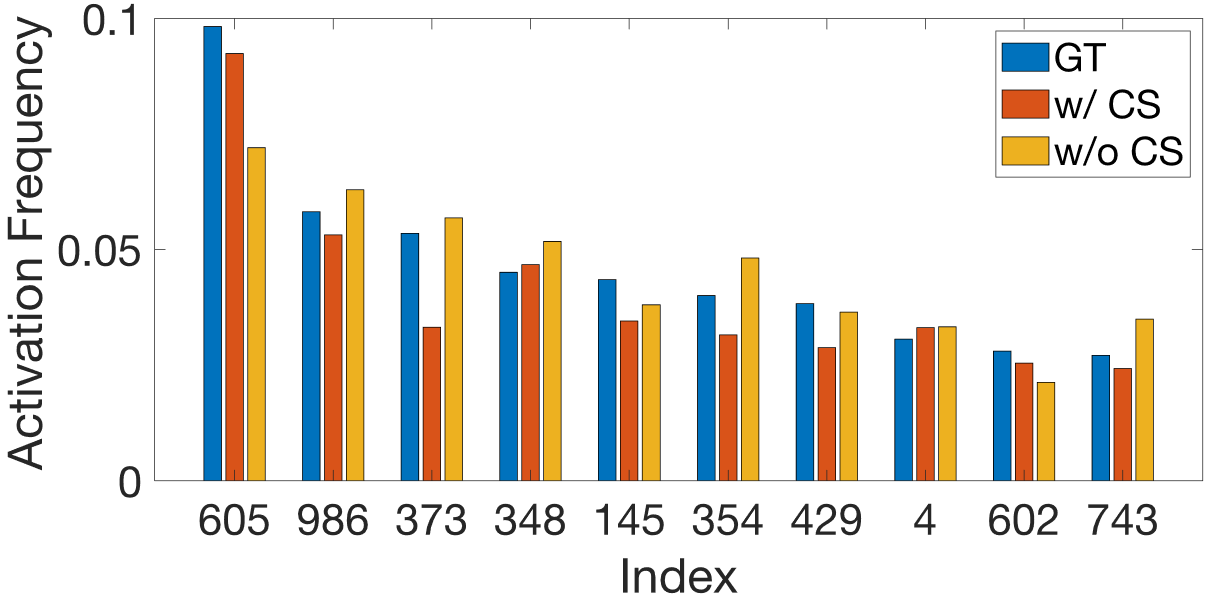}
    % \vspace{-2mm}
    \caption{{Top 10 code activation frequencies from feature quantization on the FiveK dataset \cite{FiveK}, measured under three conditions: VQ-GAN ground truth reconstruction (GT), LLIE by our model without Codebook Shift (w/o CS), and LLIE by our model with Codebook Shift (w/ CS). }}
    \label{fig:cs_ablation_img}
    \vspace{-4mm}
\end{figure}

\begin{table}[!t]
    \centering
    % \vspace{-2mm}
    \caption{{Ablation studies of the CodeEnhance on FiveK \cite{FiveK} dataset. The Baseline model is built by a VQ-GAN \cite{VQGAN}}}
    \resizebox{0.48 \textwidth}{!}{  
    \begin{tabular}{c|c|c|cc|c|cc}
    \toprule
    \multirow{2}{*}{Exp.} &\multirow{2}{*}{Baseline} & \multirow{2}{*}{SEM} & \multicolumn{2}{c|}{IFT} & \multirow{2}{*}{CS} & \multirow{2}{*}{PSNR} & \multirow{2}{*}{SSIM} \\
     &                          &                      & TFT        & CPT        &                     &                       &                       \\ \midrule
    (a) &\checkmark                         &                      &            &            &                     & 22.84                 & 0.8693                 \\
    (b) &\checkmark                         & \checkmark                    &            &            &                     & 23.17                 & 0.8845                \\
    (c) &\checkmark                         & \checkmark                    & \checkmark          &            &                     & 24.16               & 0.8955                 \\
    (d) &\checkmark                         & \checkmark                    & \checkmark          & \checkmark          &                     & 24.53                 & 0.9012                \\
 \rowcolor[HTML]{D9D9D9}   (e) &\checkmark                         & \checkmark                    & \checkmark          & \checkmark          & \checkmark                   & \textbf{24.81}               & \textbf{0.9086}                \\ 
    \bottomrule
    \end{tabular}
    } 
    \label{tab: sturcture ablation}
    % \vspace{-3mm}
\end{table}

\begin{table}[!t]
\centering
\caption{Ablation studies of IFT, reference image, and codebook on FiveK \cite{FiveK}.}
\resizebox{0.48 \textwidth}{!}{
\begin{tabular}{c|cccccc}
\toprule
Model & Baseline & Baseline + IFT & w/o Refer Img  & w/o codebook  & \cellcolor[HTML]{D9D9D9}{Ours}   \\ \midrule
PSNR  & 22.84    & 24.01          & 24.23          & 21.46 & \cellcolor[HTML]{D9D9D9}{\textbf{24.81}}  \\ 
SSIM  & 0.8693   & 0.8893         & 0.8960         & 0.8227 & \cellcolor[HTML]{D9D9D9}{\textbf{0.9086}} \\ \bottomrule
\end{tabular}}
\label{tab: sturcture ablation2}
% \vspace{-3mm}
\end{table}

\begin{table}[!t]
\centering
    \small
    \caption{{Ablation studies of the $\lambda_1$ for $\mathcal{L}_{reg}$.}}
    % \vspace{-2mm}
    \resizebox{0.35\textwidth}{!}{%
    \begin{tabular}{c|ccc >{\columncolor{gray!20}}c}
    \toprule
    $\lambda _1$ & 0      & 0.01   & 0.001  & 0.0001 \\
    \midrule
    PSNR   & 24.46  & 24.65  & 24.48  & \textbf{24.81}  \\
    SSIM   & 0.9008 & 0.9006 & 0.9014 & \textbf{0.9086} \\
    \bottomrule
    \end{tabular}}
    % \vspace{-4mm}
    % \caption{} 
    \label{tab: lambda}
\end{table}

\textbf{Study of CS.}
CS is designed to rectify distribution discrepancies among various datasets. As demonstrated in Table \ref{tab: sturcture ablation} (Exp. (d) and (e)) and Table \ref{tab: lambda}, CS has a beneficial impact on the model, resulting in an improvement of 0.16 in PSNR. 
{Additionally, Fig. \ref{fig:cs_ablation_img} illustrates the Top 10 code activation frequencies for the FiveK dataset. To measure how closely the sampling trends align with the ground truth, we use the Pearson correlation coefficient. Our results show that the model with Codebook Shift (w/ CS, p = 0.9505) exhibits a stronger alignment with the ground truth compared to the model without Codebook Shift (w/o CS, p = 0.8791), indicating that Codebook Shift effectively guides the sampling process toward the ground-truth distribution.}
% This indicates that CS can effectively fine-tune the Codebook, adapting it to the current dataset and enhancing the model's performance. 

\textbf{Study of Codebook and Reference Image.}
{Table \ref{tab: sturcture ablation2} presents an ablation study evaluating the contributions of both the reference image and the codebook. The reference image provides brightness and contrast cues that guide the enhancement process, while the codebook reframes the task as “image-to-code” and leverages high-quality priors to reduce uncertainty in the process. As shown, removing the reference image (w/o Refer Img) still produces competitive results but remains slightly below the final version, highlighting the benefit of including external brightness/contrast information. Meanwhile, omitting the codebook (w/o codebook) causes a notable performance drop, confirming that high-quality priors are critical for preserving reconstruction fidelity. }

% Finally, the full model achieves the highest PSNR (24.81) and SSIM (0.9086), indicating that combining the reference image and the codebook yields the best reconstruction quality.

\section{Conclusion}
In conclusion, we present a novel LLIE method, CodeEnhance, to obtain HQ images from LL images. We redefine LLIE as learning an image-to-code mapping between LL images and discrete codebook priors. Our method includes the Semantic Embedding Module (SEM) and Codebook Shift (CS) mechanism to enhance mapping learning by integrating semantic information and ensuring distribution consistency in feature matching. Additionally, the Interactive Feature Transformation (IFT) module improves texture and color information in image reconstruction, allowing interactive adjustment based on user preferences. Experimental results on real-world and synthetic benchmarks demonstrate that our approach improves LLIE performance in terms of quality, fidelity, and robustness to various degradations.
{However, our method is built on a GAN framework, whose image generation capability is not as strong as diffusion models. To further improve LLIE, we plan to integrate vector quantization with a diffusion model. The high-quality image representations learned by vector quantization can serve as effective prompts for the diffusion model, potentially enabling even more advanced enhancement capabilities.}
% Looking ahead, we plan to integrate the VQ with a diffusion model to further advance low-light enhancement. High-quality image representation extracted by VQ can serve as effective prompts for the diffusion model, potentially unlocking even greater enhancement capabilities.

% By leveraging VQ’s feature quantization capabilities, we can extract high-quality image representations that serve as effective prompts for the diffusion model, potentially unlocking even greater enhancement capabilities.

% \vspace{1.5em}
\noindent\textbf{Acknowledgement} \label{sec_conclusion}
This work was supported in part by the Natural Science Foundation of China (No. 61976145, No. 62076164, and No. 62272319), Guangdong Basic and Applied Basic Research Foundation (No. 2023A1515010677 and No. 2021A1515011861), and Shenzhen Science and Technology Program (No. JCYJ20210324094413037)

\ifCLASSOPTIONcaptionsoff
  \newpage
\fi

\bibliographystyle{./bib/IEEEtran}
\bibliography{./bib/main.bib}

\end{document}